\documentclass{article} 
\usepackage{colm2024_conference}

\usepackage{microtype}
\usepackage{hyperref}
\usepackage{url}
\usepackage{booktabs}
\definecolor{darkblue}{rgb}{0, 0, 0.5}
\hypersetup{colorlinks=true, citecolor=darkblue, linkcolor=darkblue, urlcolor=darkblue}

\usepackage{inconsolata}
\usepackage{adjustbox}
\usepackage{booktabs}
\usepackage{makecell}
\usepackage{natbib}
\usepackage{enumitem}
\usepackage{xspace}

\usepackage{xcolor}
\usepackage{soul}
\usepackage{multirow}

\usepackage{colortbl}
\usepackage{comment}
\usepackage{pifont}

\title{\methodemoji\method: AI-Assisted Interactive Red-Teaming for Challenging LLMs' (Lack of) Multicultural Knowledge}

\author{Yu Ying Chiu$^{1}$, Liwei Jiang$^{2,3}$,  Maria Antoniak$^{3}$, Chan Young Park$^{4}$, Shuyue Stella Li$^{2}$,\\
\textbf{Mehar Bhatia$^{5,6}$, Sahithya Ravi$^{5,6}$, Yulia Tsvetkov$^{2}$, Vered Shwartz$^{5,6}$, Yejin Choi$^{2,3}$}
\\
  $^1$ Department of Linguistics, University of Washington  \\
  $^2$ Paul G. Allen School of Computer Science \& Engineering, University of Washington \\
  $^3$ Allen Institute for Artificial Intelligence\\
  $^4$ Carnegie Mellon University\\
  $^5$ University of British Columbia\\
  $^6$ Vector Institute for AI\\ \texttt{kellycyy@uw.edu, lwjiang@cs.washington.edu}}

\colmfinalcopy 

\usepackage{float}

\newcommand*\inlineimage[1]{\raisebox{-0.15\baselineskip}{\includegraphics[height=0.98\baselineskip]{#1}$\,$}}
\newcommand{\methodemoji}{\inlineimage{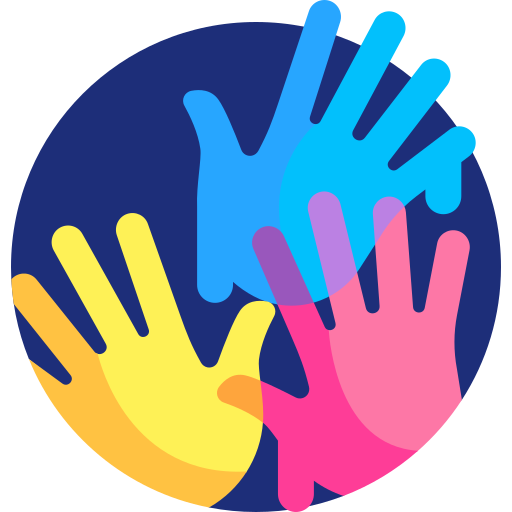}\xspace}

\newcommand{\method}{CulturalTeaming\xspace}
\newcommand{\dataset}{\textsc{CulturalBench-v0.1}\xspace}

\newcommand{\variantone}{Verifier-Only\xspace}
\newcommand{\varianttwo}{AI-Assisted\xspace}

\definecolor{stella-color}{HTML}{D687CD}

\definecolor{maria-color}{HTML}{7881F2}

\definecolor{sahithya-color}{HTML}{00308f}

\definecolor{ablation-graph-variant-1-color}{HTML}{00308f}
\definecolor{ablation-graph-variant-2-color}{HTML}{00308f}
\definecolor{workflow-variant-1-color}{HTML}{faf5e2}
\definecolor{workflow-variant-2-color}{HTML}{e4eff5}
\definecolor{chi-color}{HTML}{f5eae4}
\definecolor{european-color}{HTML}{e6f5e4}
\definecolor{gpt-3.5-color}{HTML}{fc4c4c}

\newcommand{\hlc}[2][yellow]{{%
    \colorlet{foo}{#1}%
    \sethlcolor{foo}\hl{#2}}%
}

\begin{document}
\maketitle
\begin{abstract}
Frontier large language models (LLMs) are developed by researchers and practitioners with skewed cultural backgrounds and on datasets with skewed sources. However, LLMs' (lack of) multicultural knowledge cannot be effectively assessed with current methods for developing benchmarks. Existing multicultural evaluations primarily rely on expensive and restricted human annotations or potentially outdated internet resources. Thus, they struggle to capture the intricacy, dynamics, and diversity of cultural norms. LLM-generated benchmarks are promising, yet risk propagating the same biases they are meant to measure. To synergize the creativity and expert cultural knowledge of human annotators and the scalability and standardizability of LLM-based automation, we introduce \methodemoji\method\footnote{Our platform \method is available at \url{https://cultural-norms-demo.apps.allenai.org/}.}, an interactive red-teaming system that leverages human-AI collaboration to build \textit{truly challenging} evaluation dataset for assessing the multicultural knowledge of LLMs, while improving annotators' capabilities and experiences. Our study reveals that \method's various modes of AI assistance support annotators in creating cultural questions, that modern LLMs fail at, in a \textit{gamified} manner. Importantly, the increased level of AI assistance (e.g., LLM-generated revision hints) empowers users to create more \textit{difficult} questions with enhanced perceived \textit{creativity} of themselves, shedding light on the promises of involving heavier AI assistance in modern evaluation dataset creation procedures. Through a series of 1-hour workshop sessions, we gather \dataset\footnote{Our evaluation dataset \dataset is available at \url{https://huggingface.co/datasets/kellycyy/CulturalBench}.}, a compact yet high-quality evaluation dataset with users' red-teaming attempts, that different families of modern LLMs perform with accuracy ranging from 37.7\% to 72.2\%, revealing a notable gap in LLMs' multicultural proficiency. 
\end{abstract}

\section{Introduction}

\begin{figure*}[t!]
    \centering
\resizebox{\textwidth}{!}{%
\includegraphics[width=\textwidth]{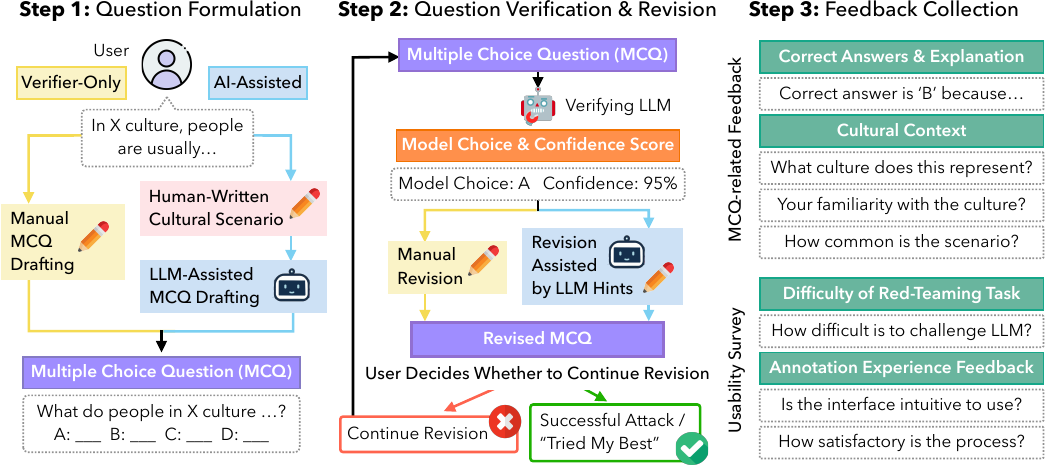}}
    \vspace{-0.4cm}
    \caption{Two settings of \methodemoji\method (1) \hlc[workflow-variant-1-color]{Verifier-Only} (2)  \hlc[workflow-variant-2-color]{AI-Assisted}. \textbf{Step 1:} Users brainstorm a culturally relevant scenario and use it to draft a multiple-choice question (MCQ). In (1), users manually draft the MCQ. In (2), an LLM drafts an MCQ based on a user-provided seed scenario. \textbf{Step 2:} Users test the question with the model and revise it iteratively until satisfied. In (1), users manually revise the MCQ. In (2), users revise with hints from an LLM. \textbf{Step 3:} Users provide gold answers and feedback.}
    \label{fig:workflow}
\end{figure*}

Uneven cultural representation has been a notorious recurrent limitation of LLMs \citep{santy2023nlpositionality,cao-etal-2023-assessing,arora-etal-2023-probing}. Yet, establishing a quality benchmark to gauge LLMs' nuanced multicultural knowledge remains a formidable challenge \citep{hershcovich2022challenges}.
Conventional human-written benchmarks are static and often fail to keep pace with the evolving capabilities of LLMs \citep{Yang2023RethinkingBA}.
Notably, crafting standardized forms of evaluation tasks, e.g., multiple-choice questions (MCQ), for such a demanding task imposes considerable cognitive and intellectual burdens on annotators. 
They must simultaneously demonstrate expert cultural knowledge, creativity in crafting novel culturally significant scenarios, familiarity of models' capabilities to pinpoint their vulnerabilities, and meticulous attention to ensure logical and format coherence within the questions.

Alternatively, automatic benchmark creation presents unique challenges. Evaluations relying on scraped online resources like bilingual content from Wikipedia \citep{Naous2023HavingBA, Yao2023EmpoweringLM}, webpages or news articles \citep{nguyen2023extracting} are susceptible to perpetuating skewed or outdated culture representations due to their limited forms and coverage of source media. Recently, established socio-cultural surveys originally designed for humans have been used to audit LLMs' cultural abilities \citep{durmus2023measuring, santurkar2023opinions}. This fails to capture the models' deficiencies. The latest approaches of creating multicultural evaluations with LLMs generations \citep{Li2024CultureLLMIC, Fung2024MassivelyMK}, despite their scalability, risk propagating existing cultural biases that they are meant to measure. It is crucial to investigate alternative data annotation strategies that harness the valuable cultural insights of humans while streamlining mundane tasks with LLMs (e.g., data formatting) to allow people to direct their cognitive resources toward substantive tasks that demand creative thinking.

Drawing insights from recent red-teaming approaches on LLMs' safety \citep{ganguli2022red, Radharapu2023AARTAR} and interactive model evaluation and data collection efforts \citep{talmor2022commonsenseqa,kiela2021dynabench, Chiang2024ChatbotAA, Zheng2023JudgingLW}, we introduce \methodemoji\method, an \textit{AI-assisted} interactive red-teaming system that guides users in iteratively creating challenging evaluation datasets for testing the multicultural proficiency of modern LLMs. To understand how effective AI assistance may boost annotators' ability to create challenging test cases, we design two variants of \method systems (Fig. \ref{fig:workflow})---\hlc[workflow-variant-1-color]{\mbox{\variantone}} that simply involves LLMs for in-the-loop testing, and \hlc[workflow-variant-2-color]{\mbox{\varianttwo}} that engages LLM assistance more intensively, including drafting and scaffolding initial MCQ candidates based on high-level topics provided by users and suggesting hints and directions during the iterative revision process.

With the two variants of \method systems, we recruited 45 annotators to participate in immersive 1-hour workshop sessions for red-teaming LLMs' (lack of) multicultural knowledge. The user study reveals positive insights about users' \textit{behaviors} and \textit{perception} during red-teaming with AI assistance. \textit{User-behavior-wise}, LLM-generated hints help users construct more challenging questions for LLMs, especially when they are depleted of ideas after three rounds of revision. In particular, with LLM-generated hints (\hlc[workflow-variant-2-color]{\mbox{\varianttwo}}), users are more willing to spend effort crafting a question with deliberation that yields more difficult questions eventually, rather than taking a rather arbitrary ``hit-or-miss'' red-teaming strategy to start over (\hlc[workflow-variant-1-color]{\mbox{\variantone}}). \textit{User-perception-wise}, users with more LLM assistance report a significantly higher level of agreement that the system sparked their \textit{creativity} than those without assistance (4.19 vs. 3.58). These human insights shed light on future data annotation practices where more AI assistance could be involved to improve annotators' ability to create high-quality data, as well as their annotating experiences.

Beyond user insights, throughout the workshop sessions, we compile users' red-teaming attempts into a compact yet high-quality MCQ dataset, namely \dataset, of 252 carefully-reviewed challenging questions spanned across 34 distinct cultures for testing LLMs' multicultural knowledge. Evaluation with \dataset reveals that \method yields difficult test cases to a variety of frontier LLM families (e.g., GPT, Claude, Mistral) across various sizes, with model performances ranging from 37.7\% to 72.2\%. This reveals a notable multicultural knowledge gap even for the most capable LLMs to date. Overall, our user study of \method demonstrates the promise of collecting high-quality, challenging datasets for multicultural knowledge by better optimizing users' cognitive resources with human-AI collaboration, laying a novel foundation for future evaluation construction to effectively assess LLMs' multicultural proficiency and beyond.

\section{\texorpdfstring{\methodemoji \method} :: Revealing the Multicultural Knowledge Gap of LLMs with AI-Assisted Interactive Red-Teaming}
\begin{figure*}[b!]
    \centering
\includegraphics[width=\textwidth]{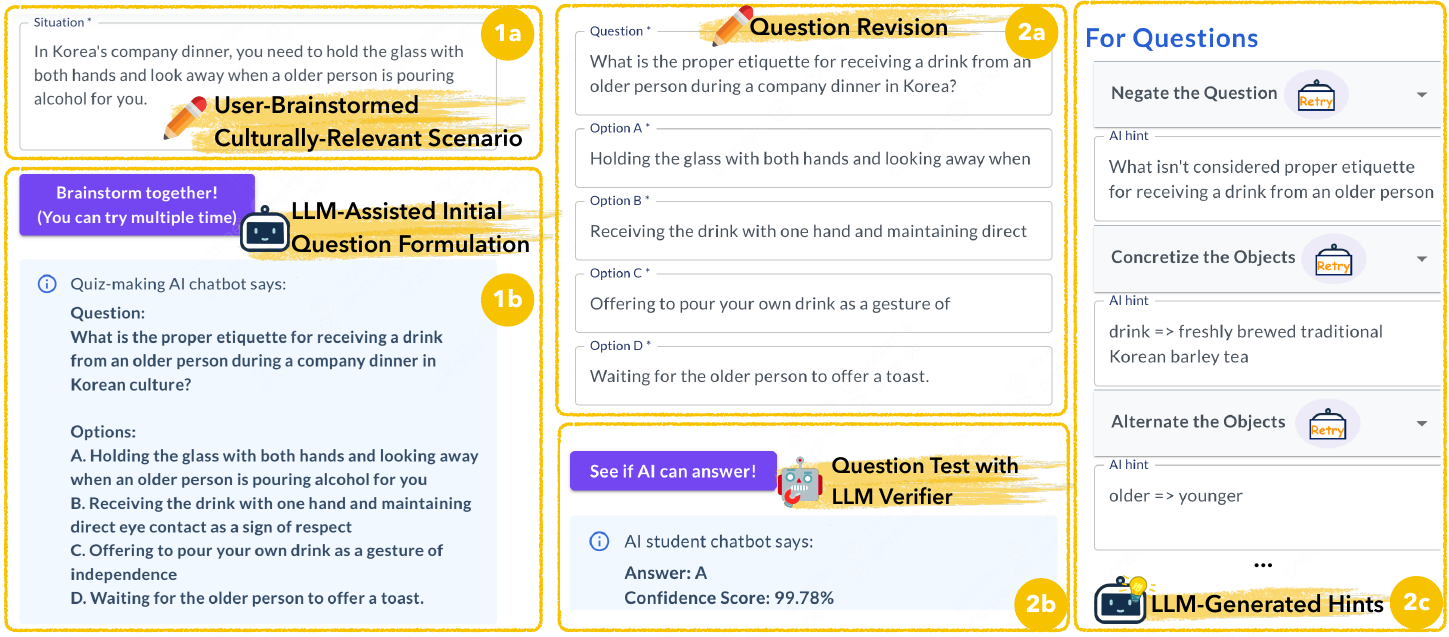}
\vspace{-0.4cm}
    \caption{\method Interface: \textbf{(1a)} Users brainstorm culturally relevant scenarios \textbf{(1b)} They convert scenarios to MCQs with LLM-powered Question Formulation \textbf{(2a)} Users revise MCQs and \textbf{(2b)} test MCQs based on the chosen option and its confidence score from LLM Verifier \textbf{(2c)} Users inspire by LLM-generated hints with strategies e.g., Negation, Synonym.}
    \label{fig:modules}
\end{figure*}
Our goal is to build a data annotation platform to help human annotators create a challenging and structured dataset based on their expert cultural knowledge.
\subsection{Overall System Workflow}
This system consists of three steps, as demonstrated in Fig. \ref{fig:modules} -- 1) Question Formulation 2) Question Verification and Revision 3) Feedback Collection. The first two steps involve a red-teaming exercise to formulate a challenging question step-by-step. The last step involves collecting feedback from users through a cultural contextualization survey based on their posed questions, as well as a usability survey regarding their user experience. 
\textbf{Step 1: Question Formulation}
The goal is to facilitate users in brainstorming culturally relevant situations based on their personal experiences. A step-by-step guideline with detailed examples is provided to inspire them, as shown in Appendix \ref{fig:guidance-step-1}. Users formulate a multiple-choice question (MCQ), which comprises one correct and culturally appropriate option.
\textbf{Step 2: Question Verification \& Revision}
This step provides an interactive and iterative red-teaming platform that allows users to verify their culturally sensitive MCQs. The platform assists them in revising the question and the options to make it more challenging by providing descriptions of various common revision strategies with drafted examples (e.g., ``Negate the Question''), as stated in Fig. \ref{fig:modules} and Appendix  \ref{fig:guidance-step-2}. 
\textbf{Step 3: Feedback Collection}
After finalizing a question, we collected users' feedback on the question posed using a cultural contextualization survey and also their user experience by usability survey as depicted in Fig. \ref{fig:modules}.
The cultural context survey asked users to provide their gold standard (correct option), its rationale, and several questions about their levels of particular cultural knowledge and the commonness of the specific cultural norm described in the question. We also gathered their hands-on experience regarding the functionality and user-friendliness of our interface. Survey questions can be found in Appendix \ref{app:survey_questions}.
\subsection{Varying Levels of AI Assistance: \variantone \& \varianttwo}
To study the effect of different AI-assistance modules, we have implemented two settings with varying AI Assistance levels in our experiment: (1) \variantone (2) \varianttwo, as presented in Fig. \ref{fig:workflow}. The first setting includes the basic element (Verifier) for the red-teaming exercise. The second setting aims to explore question formulation and hints on revision powered by LLMs. 
\textbf{\variantone} The users formulate their MCQ in Step 1 (Question formulation). They then present their MCQ to LLM Verifier and revise their question iteratively by themselves. More specifically, the Verifier responds with one option coupled with the corresponding confidence score (which is the top-1 linear probability of its log probability). This aids users in judging the question's difficulty. In our system, GPT-3.5-turbo (0125) served as the Verifier to reduce cost and time latency  \citep{brown2020language}.
\textbf{\varianttwo} Users benefit from extensive AI-assistance in question formation and revision, as illustrated in Fig. \ref{fig:modules}. In Step 1, users convert their short cultural-relevant scenarios into MCQs using LLM-powered question formation. In Step 2, users verify and revise their questions with LLM-powered suggestions. The Verifier remains consistent with the \variantone setting.

\subsection{User Experiments with Red-teaming Workshops}
We recruit annotators from an academic institution to participate in our 1-hour workshop. Annotators are randomly assigned to use either \variantone or \varianttwo. After finishing the self-contained tutorial, we encourage annotators to interact with the system for at least 45 minutes, allowing them to leave at their own will. Additional details regarding the user demographic description, and experiment setup can be found in Appendix \ref{sec:appendix-user-experiment} and \ref{app:demographic}.
\subsection{\dataset: A Compact Dataset for LLMs' Multicultural Knowledge}
\label{sec:dataset_model_eval}

\begin{table*}[!ht]
\centering
\resizebox{\textwidth}{!}{
\begin{tabular}{l|l|l|l} 
 \toprule
 & \textbf{Question} & \textbf{Models} & \textbf{Rationale} \\
 \midrule
\makecell[c]{(1) \\ Indian} & \makecell[l]{\textbf{Q: What is defined as vegetarian in India?}\\A) Not eating seafood, meat, or eggs\\B) Not eating seafood or meat\\C) Not eating seafood, meat, eggs, or root vegetables\\D) All of the above}  & \makecell[l]{GPT-4: D \ding{51}\\Others: A, B} & \makecell[l]{All of the above \\  are considered \\  vegetarian in  \\ India, but just \\ different types.}\\
\midrule
\makecell[c]{(2) \\ South \\ Korean}  & \makecell[l]{\textbf{Q: What are some unspoken etiquette in a company dinner in Korea?}\\A) Hold the glass with both hands\\when a younger person is pouring alcohol for you\\B) Look away from elders while drinking alcohol\\C) Maintain eye contact the entire time\\D) Make sure to start eating first} & \makecell[l]{GPT-4: B \ding{51}\\ Others: A} & \makecell[l]{For A) It is only \\ true when the \\ person pouring \\ the alcohol is \\ older than \\ you...} \\

\midrule
\makecell[c]{(3) \\ Chinese} & \makecell[l]{\textbf{Q: In a Chinese cultural context, what will the government} \\ \textbf{do if a couple gives birth to a second child?} \\A) Government issue a fixed amount of fine because this is not allowed.\\B) Government issue a variable amount of fine based on\\the couple's income level because this is not allowed\\C) Government does not give cash bonus or cash punishment.\\D) Government sends cash bonus to improve the low birth rate in China.} & \makecell[l]{GPT-4: C \ding{51} \\ Yi: C \ding{51} \\ Others: A, B} & \makecell[l]{Having a \\ second child \\ is allowed in \\ China after \\ 2014.} \\
\bottomrule
\end{tabular}}
\caption{Examples from \dataset and models responses. The detailed rationale can be found in Appendix \ref{table:error_models}.}
\label{table:examples_dataset}
\end{table*}

Before delving into the insights of the user studies, we first introduce the end dataset that we gathered by \method. We collected not only question-specific information (Fig. \ref{fig:dataset_schema}-Left e.g., Correct answer, rationale), but also cultural contextualization details (Fig. \ref{fig:dataset_schema}-Right e.g., culture represented, annotator's familiarity on it). Both success-attack questions and fail-to-attack questions have the same metadata, as shown in Appendix \ref{app:success_attack} and \ref{app:fail_attack}.

Our red teaming dataset includes 252 carefully-reviewed MCQs spanned across 34 diverse cultures. After data collection, we manually reviewed the questions to ensure they 1) follow the MCQ format 2) specify what kind of culture the question is asking for.

This dataset represents diverse demographic groups as stated in Appendix \ref{app:demographic}. Chinese culture, including ethnicity Chinese in different locations, attributes the highest proportion ($32\%$) among all, following are Indian (including North and South Indian) and American containing $15\%$ and $6\%$ respectively. We also include Appendix Table \ref{table:dataset-culture-represent-example} to show the grouping examples for some cultures with confusing terms (e.g., South Indian and Hindu Indian are grouped as Indian).

\section{Promising AI-Assisted Benchmark Creation with \method}

This section aims to provide insights on how to make use of human-AI collaboration to produce a challenging dataset on multicultural knowledge. 

We analyze (1) AI-assistance contribution to different system modules, (2) existing LLMs' cultural awareness using \dataset, and (3) behavior and perception of users with varying levels of AI-assistance.

\subsection{How does AI-assistance contribute to each module in our system?}

We study each LLM-powered module to investigate the effect of human-AI collaboration on each type of task in our \method system.

\subsubsection{LLM assistance on question formulation}
Question formulation entails brainstorming a culturally relevant scenario and transforming it into a multiple choice question (MCQ). Users with LLM-powered question formulation are only tasked to provide a sentence-length scenario description, whereas users without LLM-assistance need to formulate the MCQ question and answer options themselves. We first compare the time difference in the question formulation between the two groups.

\textbf{LLM assistance on question formulation \textit{does not significantly improve} the time needed on formulating questions.}
The time needed to form the initial MCQ template before editing for both users with more AI-assistance on question formulation (\varianttwo setting) and less AI-assistance (\variantone setting) are shown in Appendix \ref{app:result-time-needed}. The time needed for the initial MC questions created by \variantone and \varianttwo settings show no significant difference. Specifically, the former needs 183.4 seconds (SD=105.2) on average while the latter needs 161.8 (SD=109.4) on average.

Our formulation of red teaming tasks by writing structured MCQs appears to be relatively easy for human users without AI assistance. They effectively create their initial question template and attain equally impressive performance in terms of efficiency. One possible reason could be users without AI assistance formulate both questions and options concurrently when they are brainstorming. As we move forward in our future work, we should also consider how AI can aid people in brainstorming culturally relevant scenarios, rather than solely helping them in formulating structured questions.

\begin{figure}[h!]
    \begin{minipage}{0.45\textwidth}
        \centering
        \includegraphics[width=\textwidth]{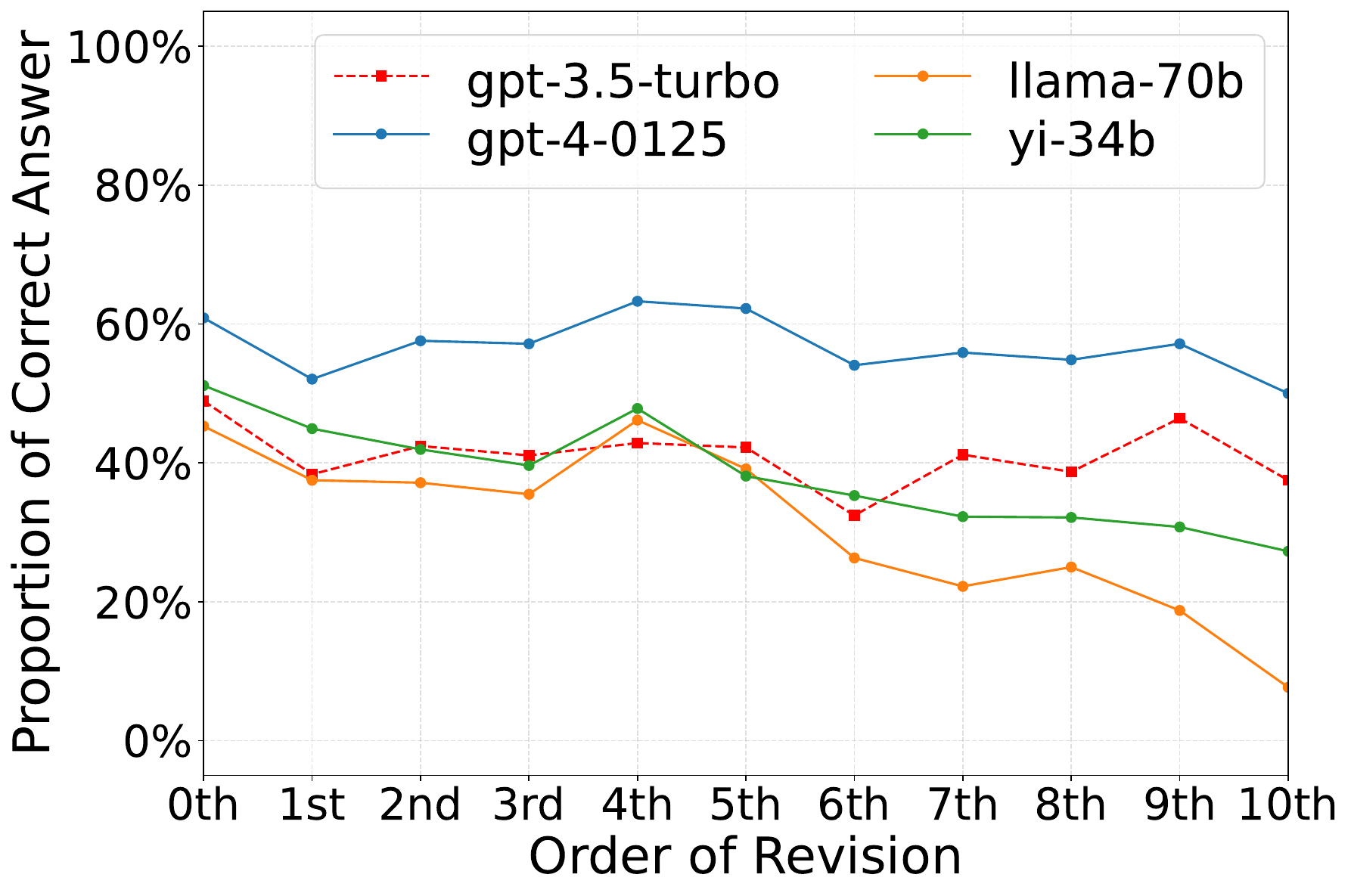} 
    \end{minipage}
    \begin{minipage}{0.45\textwidth}
        \centering
\includegraphics[width=\textwidth]{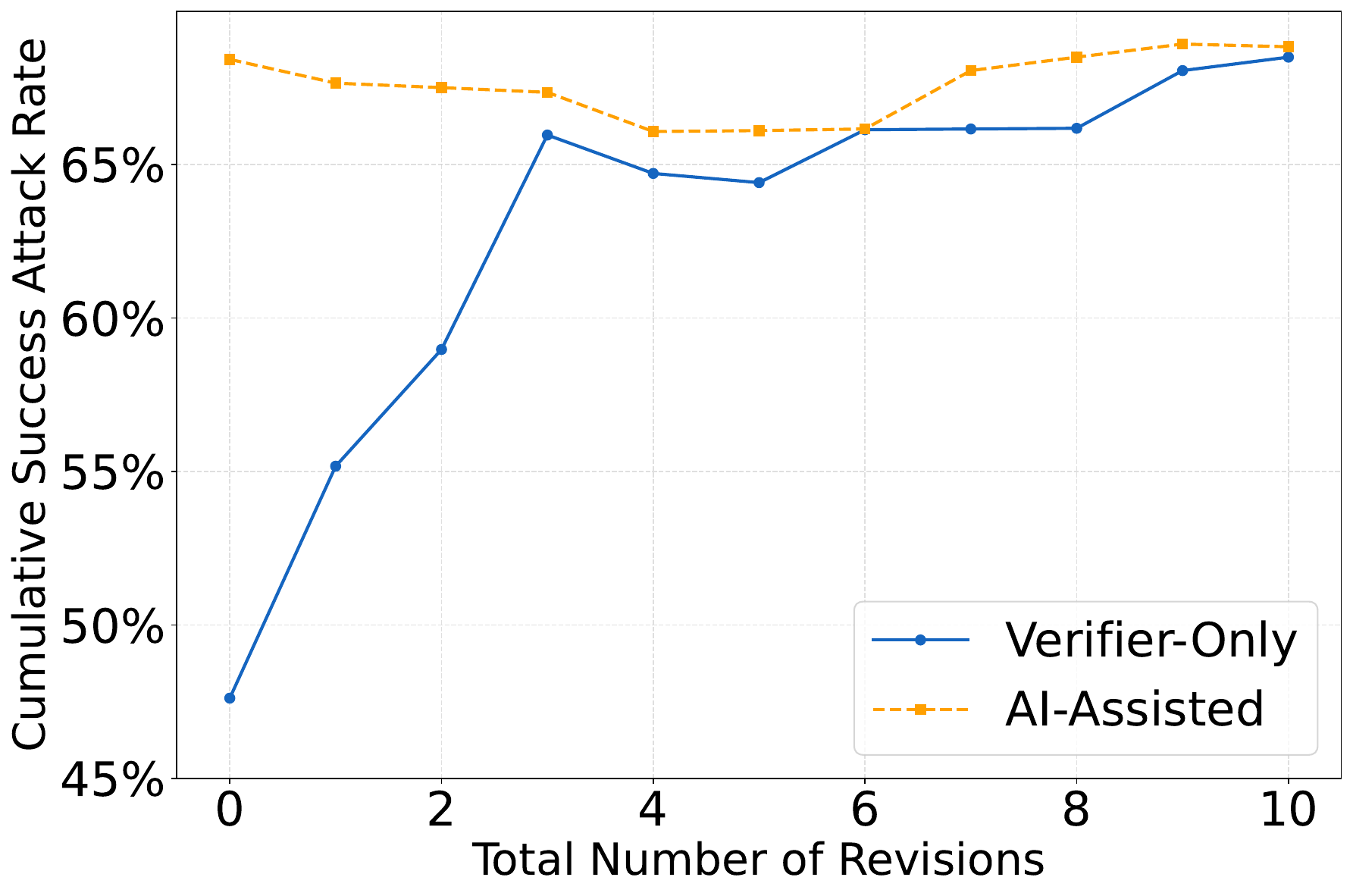}
    \end{minipage}
\caption{LLM assistance on (1) \textbf{Left}: Verifier (\textcolor[HTML]{fc4c4c}{gpt-3.5-turbo}) by comparing other models performance on questions by users without other LLM assistance (\variantone) (2) \textbf{Right}: Revision by comparing between the final success attack rate and the total number of edits between users with LLM Hints (\textcolor[HTML]{f4800b}{\varianttwo}) and without LLM Hints (\textcolor[HTML]{404ba4}{\variantone}).}
\label{fig:combined-ablation-graphs}
\end{figure}

\subsubsection{LLM assistance on Question Verifier}
After the MCQ is initially formulated, we attempt to use GPT-3.5-turbo as our Verifier \citep{brown2020language}. Users revise and verify the MCQ iteratively based on the Verifier response.
In this section, we evaluate the viability of this setup by investigating whether the Verifier serves as a good estimator for the degree of challenge for other LLMs. We assume that users revise their MCQs based on the Verifier response and its corresponding confidence score.

\textbf{Our LLM Verifier \textit{is able to create challenging questions} for all tested models.} To delineate the effect of LLM assistance, we analyze the MCQs created by users without LLM assistance on other LLMs in Fig. \ref{fig:combined-ablation-graphs} (Left). We found that LLM performance decreases with the number of revisions, especially for llama-70b by 38.73\% and yi-34b by 23.82\%. GPT-3.5-turbo (our Verifier model) decreased from 48.91\% to 37.5\%. GPT-4-turbo, which is the best performance model among all, decreased from 60.87\% to 50.00\% on the 10th revision. This shows that our LLM Verifier is capable of increasing the difficulty of questions for most models, as the number of revisions increases based on the response of LLM Verifier.

It implies that using only one model (GPT-3.5-turbo) as the Verifier can effectively construct challenging data for most of the models. However, if the goal of future work is to improve the stronger model (GPT-4-turbo), the relatively small decrease in it suggests that we need to randomly select different models as verifier, including GPT-4 model to help constructing a more challenging dataset.

\subsubsection{LLM assistance on Question Revision}
GPT-4 is used to generate LLM-powered hints based on the work-in-progress MCQ to provide revision suggestions with common revision strategies. Users without LLM-powered hints, on the other hand, are provided with static descriptions of those strategies. We now investigate the relationship between the number of revisions and the success attack rate of the finalized question (final success rates) for both users with LLM-assistance (\varianttwo setting) and users without LLM-assistance (\variantone settings).

\textbf{LLM assistance on Hints helps users to construct \textit{more challenging questions} after several rounds of revision.} 
For the initial three revisions, the final success attack rate of questions without LLM-powered suggestions (\variantone) increases. Conversely, questions with LLM-powered suggestions (\varianttwo) remain steady as they are initially created by the LLM-powered question formulation and are of good quality. Beyond four revisions, questions created in \varianttwo begin to show an increasing trend. On the other hand, those in \variantone require more revisions to match similar performance at 10 revisions, relative to those in \varianttwo. We can see the potential benefits of LLM suggestions become more apparent, especially after several rounds of revisions.

One reason is that individuals start with their initial ideas, which may deplete after a few attempts. LLM suggestions can therefore serve as an idea pool, providing diverse editing ideas that are grounded in their input question template. This illustrates the use of AI in facilitating creative thought and efficient question revision among individuals.

\subsection{Assessing LLMs' Cultural Knowledge Gap with \dataset}
\label{sec:dataset-analysis}

\begin{table*}[t]
    \centering
    \resizebox{\textwidth}{!}{%
    \begin{tabular}{l|c|c|c|c|c|c|c} 
     \toprule
\textbf{Model}&\makecell{\textbf{All}\\(N=252)}&\makecell{\textbf{Verifier-Only}\\(N=113)}&\makecell{\textbf{AI-Assisted}\\(N=139)}&\makecell{\textbf{Chinese}\\(N=80)}&\makecell{\textbf{Indian}\\(N=38)}&\makecell{\textbf{American}\\(N=16)}&\makecell{\textbf{Europe}\\(N=46)}\\
\midrule
Claude 3 Opus (20240229) & 70.6 & 69.9 & 71.2 & 66.2 &\textbf{71.1} & \textbf{81.2} & \textbf{69.6}\\
Claude 3 Sonnet (20240229) & 57.1 & 61.9 & 53.2 & 50.0 & 57.9 & 75.0 & 60.9\\
GPT-4 Turbo (0125) & \textbf{72.2} & \textbf{71.7} & \textbf{72.7} & \textbf{67.5} & \textbf{71.1} & \textbf{81.2}& \textbf{69.6}\\
GPT-3.5 Turbo (0125) & 39.7 & 41.6 & 38.1 & 41.2 & 44.7 & 56.2 & 41.3\\
Mistral Large & 57.1 & 62.8 & 52.5 & 55.0 & 57.9 & 81.2 & 54.3\\
Mixtral MoE model (8x7b) & 50.4 & 51.3 & 49.6 & 45.0 & 47.4 & 75.0 & 52.2\\
Llama (70b) & 44.4 & 42.5 & 46.0 & 41.2 & 47.4 & 43.8 & 50.0\\
Mistral (7b) & 37.7 & 38.1 & 37.4 & 40.0 & 39.5 & 68.8 & 26.1\\
Yi (34b) & 44.8 & 48.7 & 41.7 & 41.2 & 50.0 & 68.8 & 37.0\\
\midrule

\textbf{Average} &$ 52.69_{11.89}$ &$ 54.28_{11.94}$ &$ 51.4_{12.23}$ &$ 49.72_{10.27}$ &$ 54.09_{10.61}$ &$ 70.14_{12.07}$ &$ 51.21_{13.79}$\\
    \bottomrule
    \end{tabular}}
    \caption{\dataset: Zero-shot results (correct answer rate in \%) by greedy decoding from LLMs in terms of culture represented (Chinese, Indian, American) and geographical locations (Europe) by grouping cultures e.g. French, Italian. 
    }
    \label{table:other-llms-results-combined}
 \end{table*}

With our high-quality and small-scale dataset, we evaluate several state-of-the-art LLMs to assess their cultural awareness and conduct qualitative analysis on factors that contribute to the difficulty of questions for LLMs. 

In Table \ref{table:other-llms-results-combined}, we present the performance of several open-source models (e.g., Mistral, Yi) and state-of-the-art private models from different model families (e.g., Claude variants, GPT variants). Additional evaluation details can be found in Appendix \ref{sec:appendix-evaluation-detail}.

\textbf{\dataset reflects cultural-specific knowledge \textit{with mixed results} for the Chinese model (Yi) and European models (Mistral/Mixtral) on their relevant cultures.} Contrary to the intuitive belief that models trained with a special focus on specific languages (e.g., Yi for Chinese and Mistral large and Mixtral MoE models for European languages) will perform better on cultures associated with the specialized language, we surprisingly observe mixed results. While Yi demonstrates lower-than-average performance (41.00 vs. 49.72), Mistral Large and Mixtral MoE (54.3 and 52.2 vs. 51.21) show significantly higher performance than average.

As illustrated in Table 3, for example, only Yi and GPT-4 models could answer correctly on the updated one-child policy in China: ``\textit{In a Chinese (...), what will the government do if a couple give birth to a second child?}'' when other models still believed only one child is allowed in China. However, Yi fails to give a correct answer to a question about driving licenses: ``\textit{How many tests do people have to pass to get a driver's license in China?}'' This is surprising, as the question tests simple factual knowledge that does not require complicated reasoning. 
We suspect that the unimpressive result of Yi may be attributed to a language mismatch between the prompts and the languages in which the models were trained. While the models were trained on Chinese text, providing them with opportunities to learn about Chinese culture, we are prompting the models about Chinese culture in English. Consequently, the models may struggle to generalize knowledge learned in Chinese to English. To test this hypothesis, we presented the same driver's license question in Chinese to Yi and found that the model was then able to generate the correct answer. 
Additional evaluation details and model response can be found in Appendix \ref{sec:appendix-evaluation-detail}. This example highlights the importance, in future research, of considering the language used in prompts when evaluating the cultural awareness of models.

\textbf{\dataset is \textit{challenging to modern LLMs}, with varying difficulties.}

Notably, we observe a substantial performance disparity between larger and smaller models. For example, GPT-4-turbo achieves the highest accuracy at 72.2\%, closely followed by Claude 3 Opus at 70.6\%. Conversely, relatively smaller models such as Mistral 7b and GPT-3.5-turbo attain lower accuracy scores of 37.7\% and 39.7\%, respectively. 

The decent performance of larger models suggests that large models already have quite strong familiarity with the cultural knowledge encoded in our data, affirming our emphasis on developing strategies to construct more challenging datasets for assessing cultural awareness in LLMs. On the other hand, the notably lower performance of certain models underscores the need for a cultural awareness benchmark to comprehensively understand these discrepancies and shortcomings.

\textbf{\textit{Hard} questions in \dataset require \textit{more extensive reasoning}.} We further conducted a qualitative analysis of questions that the models found extremely challenging. We categorized our dataset into three difficulty levels based on the number of models that provided correct answers: \textit{Easy} (7-9 models, 68 samples), \textit{Medium} (3-6 models, 103 samples), and \textit{Hard} (0-2 models, 81 samples). We then analyzed the \textit{Hard} questions, focusing specifically on those that only GPT-4 could answer correctly. Examples of such questions are provided in Table \ref{table:examples_dataset} (Example 1-3) and Appendix \ref{app:detail-eg-models}.

Our analysis reveals that \textit{Hard} questions often demand more extensive reasoning from the models due to the complexity of the question structure and the presence of subtly incorrect options. For instance in Table \ref{table:examples_dataset} (Example 1), ``\textit{What is defined as vegetarian in India?}" with the correct answer being ``\textit{D. All of the above.}" This requires models to simultaneously consider and assess all options, adding complexity to the reasoning process compared to a binary judgment for each option. Another example is the question in Table \ref{table:examples_dataset} (Example 2), ``\textit{What are some unspoken etiquettes during company dinners in Korea?}" where only GPT-4 identified the correct response, ``\textit{Look away from elders while drinking alcohol}." Other models were misled by selecting ``\textit{holding the glass with both hands when a younger person is pouring alcohol for you}," which is incorrect as it applies to the ``older" person, not the ``younger." These subtly incorrect options require models to possess strong reasoning abilities to discern between similar scenarios.

\subsection{How users' behaviors and perception differ with varying levels of AI-assistance}

We compare and analyze annotators' behaviors and their perception. We used a two-sided Student t-test for the following analysis.

\subsubsection{Behaviors}
\label{sec:user-behavior}

\textbf{Users with more AI-assistance spend \textit{more time} on revising and make \textit{more revisions}, compared with those with less AI-assistance.} 
As shown in Table \ref{table:results-time-taken-combined}, users with LLM-generated hints (\varianttwo setting) require 152.4 more seconds, compared to users without LLM-generated Hints  (\variantone) with $p=0.1$. 
Similarly, users in \varianttwo setting also make 7.19 more revisions than users in \variantone. One possible reason is that the LLM-generated hints provide more information and inspiration for users to revise their questions. Users with LLM-generated hints tend to try and adopt various existing hints whereas users without LLM-generated hints tend to give up on revising and restart another round due to a lack of revision ideas and less guidance.

\textbf{User with more AI-assistance write \textit{more Hard questions}, relative to users with less AI-assistance.} 
Following the Section \ref{sec:dataset-analysis} questions categorization,  we found that each user with \varianttwo created 15.7\% more \textit{Hard} questions (0-1 models we selected in Table \ref{table:other-llms-results-combined} correct correctly) more than each user without \varianttwo at p=0.05. 

This finding could indicate that revision based on posed questions attributed to the creation of more Hard questions for LLMs. 
This demonstrates that encouraging users to continue revising rather than restarting a new scenario could be a key to creating challenging LLM benchmarks.

\subsubsection{Perception}

\textbf{Users with more AI-assistance expressed \textit{more positive agreement on the system capability to spark their creativity}, relative to those with less AI-assistance.}
On a scale of 1 (very limiting to creativity) to 5 (very conductive to creativity), users with more AI-assistance (\varianttwo setting) reported a mean of 4.19 (SD=0.66) whereas users with less AI-assistance (\variantone) reported a mean of 3.58 (SD=0.79). 
The difference between these scores (0.61) is significant ($p=0.05$). 
This suggests that the inclusion of LLMs in the initial question forming or revision could also potentially boost the system's capacity to stimulate creativity. 
One user expressed similar thoughts: ``\textit{The ability to generate multiple iterations of questions using AI was helpful to my creativity, as well as the hints provided.}'' 

\textbf{\textit{Nine} users with both more and less AI-assistance report \textit{positive impressions} of the annotation system and task design.} 

9 out of 27 users (5 from \variantone; 4 from \varianttwo) expressed their enthusiasm. One user from \variantone said ``\textit{I enjoyed learning how much the AI actually knew about culture,}''  Another user from \varianttwo liked the gamified elements: ``\textit{I liked how we could run the model again and again, ..., making this process a lot more fun and gamified.}''
Users from both settings positively referred to ``the confidence score'' helping them understand the effectiveness of their revisions.

\textbf{\textit{Some} users with both more and less AI-assistance \textit{enjoy} the LLM-powered modules.} 
Some users appreciate knowing the confidence score along with the chosen option by our Verifier (3 from \variantone, 1 from \varianttwo). This allows them to gauge how close they are to 'winning' (trick the AI successfully). Users with more AI-Assistance (6 from \varianttwo) enjoyed various LLM-powered assistance, emphasizing ``\textit{I liked how it generated a question based on your observation of a culture}''(4 from \varianttwo) and ``\textit{AI-generated hints} (2 from \varianttwo).''

One possible reason is that the LLM-powered modules reduce users' cognitive load when formulating questions and revisions. 
They provide grounded ideas based on users' inputs (e.g. their scenario, their posed question), rather than purely text guidance. Users engaged more by trying the LLM-powered modules, possibly increasing question revision time. 
When designing similar annotation systems, the trade-off between engagement and time spent should be considered.

\textbf{\textit{Six} users with both more and less AI-assistance report \textit{difficulty in deceiving the LLM}.} 
11 out of 27 users (6 from \variantone; 5 from \varianttwo) believed that nothing posed a challenge.  
However, 6 users (2 from \variantone; 4 from \varianttwo) expressed difficulty in deceiving the LLM: ``\textit{Refine the questions was a bit challenging. That might be due to the nature of the question rather than the platform itself.}''. Despite all our design modifications, such as incorporating a confidence score along with the response option, this remains difficult for some users, as evidenced by their feedback. This underscores the importance of designing novel annotation systems to assist human annotators in brainstorming scenarios rather than purely formulating questions.

\section{Related Work}
\textbf{AI-augmented Data Collection} 
Prior works have investigated on how to incorporate AI into different subtasks during data curation including writing \citep{shen2023convxai} and auto-crowdsourcing pipelines \citep{wu2023llms}. One popular approach to improve LLMs is through red teaming to reveal models' weaknesses. Researchers recruited human annotators to attack their LLMs to reduce models' harmful outputs \citep{ganguli2022red}. Some works have also been done on constructing adversarial datasets more efficiently with AI massive generations \citep{Radharapu2023AARTAR}.

\textbf{Cultural Bias on LLMs} To evaluate using grounded theory, some studies focus on auditing LLMs and comparing them to humans through existing socio-cultural surveys \citep{Li2024CultureLLMIC, Tao2023AuditingAM, ramezani2023knowledge}. People also investigate language diversity by scaling up to multilingual settings \citep{Mukherjee2023GlobalVL, yue2023mmmu}, with a particular focus on low-resource languages \citep{Deas2023EvaluationOA, Wibowo2023COPALIDIL}. Some recent works have also been conducted to construct schemas for generating massive datasets with AI, which aims to better capture LLMs' cultural bias \citep{Fung2024MassivelyMK, huang2023culturally, ziems2023normbank}. 
\section{Conclusion}
We present \method, an interactive red-teaming system that leverages the synergy of human-AI collaboration to collect a truly challenging dataset to assess LLMs' multicultural knowledge. Through workshop sessions in our user studies, we gather users' red-teaming attempts to form a compact yet high-quality evaluation dataset \dataset. With varying AI-assistance levels of our system, we compared and analyzed our user studies' results in terms of system performance, data quality, and users' behaviors and perception.
\section*{Acknowledgements}
We thank the users of our \method platform for their helpful feedback and comments. This research was supported in part by DARPA under the ITM program (FA8650-23-C-7316).
\newpage
\section*{Ethics Statement}

Our study recruited students from a Natural Language Processing class at an academic institution. We provided participation class credits to participants who joined the experiment. These credits can also be gained through other methods e.g., writing essays. Therefore, the participation is optional. Before joining the workshop, 
we provide consent materials to explain the experiment's setup and their rights, including the ability to ask for clarification anytime and leave at their own will during the experiment. This study was reviewed by the same academic institution's IRB committee.

The findings of the user studies on our \method system are based on this limited user demographics and designed task guidance. Our users were mostly from the computer science department and may not represent the general public, as shown by the skewed population in Table \ref{table:demographic_overall}. Additionally, our users were more familiar with interacting with LLM-based systems when compared with the general public. In this study, we prepared detailed guidance and tutorials for users to do their red-teaming tasks. This approach may not be natural, and we would need to consider such differences when employing our system with other public user groups in the future. 

Moreover, the \dataset is a small-scale dataset. Although it covers a wide range of cultures, many of them only contain a few data points. We have also chosen to collect English datasets this time instead of multilingual datasets. Therefore, the dataset is not sufficient to evaluate the multicultural knowledge of LLMs. However, we believe that \method has huge potential of collecting various cultural norms by expanding this platform to different locations and languages in the future.
\newpage
\bibliography{colm2024_conference}

\begin{thebibliography}{41}
\providecommand{\natexlab}[1]{#1}
\providecommand{\url}[1]{\texttt{#1}}
\expandafter\ifx\csname urlstyle\endcsname\relax
  \providecommand{\doi}[1]{doi: #1}\else
  \providecommand{\doi}{doi: \begingroup \urlstyle{rm}\Url}\fi

\bibitem[Anthropic(2024)]{TheC3}
Anthropic.
\newblock The claude 3 model family: Opus, sonnet, haiku.
\newblock In \emph{Anthropic Model Card}, 2024.
\newblock URL \url{https://www-cdn.anthropic.com/de8ba9b01c9ab7cbabf5c33b80b7bbc618857627/Model_Card_Claude_3.pdf}.

\bibitem[Arora et~al.(2023)Arora, Kaffee, and Augenstein]{arora-etal-2023-probing}
Arnav Arora, Lucie-aim{\'e}e Kaffee, and Isabelle Augenstein.
\newblock Probing pre-trained language models for cross-cultural differences in values.
\newblock In Sunipa Dev, Vinodkumar Prabhakaran, David Adelani, Dirk Hovy, and Luciana Benotti (eds.), \emph{Proceedings of the First Workshop on Cross-Cultural Considerations in NLP (C3NLP)}, pp.\  114--130, Dubrovnik, Croatia, May 2023. Association for Computational Linguistics.
\newblock \doi{10.18653/v1/2023.c3nlp-1.12}.
\newblock URL \url{https://aclanthology.org/2023.c3nlp-1.12}.

\bibitem[Brown et~al.(2020)Brown, Mann, Ryder, Subbiah, Kaplan, Dhariwal, Neelakantan, Shyam, Sastry, Askell, Agarwal, Herbert-Voss, Krueger, Henighan, Child, Ramesh, Ziegler, Wu, Winter, Hesse, Chen, Sigler, Litwin, Gray, Chess, Clark, Berner, McCandlish, Radford, Sutskever, and Amodei]{brown2020language}
Tom~B. Brown, Benjamin Mann, Nick Ryder, Melanie Subbiah, Jared Kaplan, Prafulla Dhariwal, Arvind Neelakantan, Pranav Shyam, Girish Sastry, Amanda Askell, Sandhini Agarwal, Ariel Herbert-Voss, Gretchen Krueger, Tom Henighan, Rewon Child, Aditya Ramesh, Daniel~M. Ziegler, Jeffrey Wu, Clemens Winter, Christopher Hesse, Mark Chen, Eric Sigler, Mateusz Litwin, Scott Gray, Benjamin Chess, Jack Clark, Christopher Berner, Sam McCandlish, Alec Radford, Ilya Sutskever, and Dario Amodei.
\newblock Language models are few-shot learners, 2020.

\bibitem[Cao et~al.(2023)Cao, Zhou, Lee, Cabello, Chen, and Hershcovich]{cao-etal-2023-assessing}
Yong Cao, Li~Zhou, Seolhwa Lee, Laura Cabello, Min Chen, and Daniel Hershcovich.
\newblock Assessing cross-cultural alignment between {C}hat{GPT} and human societies: An empirical study.
\newblock In Sunipa Dev, Vinodkumar Prabhakaran, David Adelani, Dirk Hovy, and Luciana Benotti (eds.), \emph{Proceedings of the First Workshop on Cross-Cultural Considerations in NLP (C3NLP)}, pp.\  53--67, Dubrovnik, Croatia, May 2023. Association for Computational Linguistics.
\newblock \doi{10.18653/v1/2023.c3nlp-1.7}.
\newblock URL \url{https://aclanthology.org/2023.c3nlp-1.7}.

\bibitem[Chiang et~al.(2024)Chiang, Zheng, Sheng, Angelopoulos, Li, Li, Zhang, Zhu, Jordan, Gonzalez, et~al.]{Chiang2024ChatbotAA}
Wei-Lin Chiang, Lianmin Zheng, Ying Sheng, Anastasios~Nikolas Angelopoulos, Tianle Li, Dacheng Li, Hao Zhang, Banghua Zhu, Michael Jordan, Joseph~E Gonzalez, et~al.
\newblock Chatbot arena: An open platform for evaluating llms by human preference.
\newblock \emph{arXiv preprint arXiv:2403.04132}, 2024.

\bibitem[Clark et~al.(2020)Clark, Choi, Collins, Garrette, Kwiatkowski, Nikolaev, and Palomaki]{Clark2020TyDiQA}
J.~Clark, Eunsol Choi, Michael Collins, Dan Garrette, Tom Kwiatkowski, Vitaly Nikolaev, and Jennimaria Palomaki.
\newblock Tydi qa: A benchmark for information-seeking question answering in typologically diverse languages.
\newblock \emph{Transactions of the Association for Computational Linguistics}, 8:\penalty0 454--470, 2020.
\newblock URL \url{https://api.semanticscholar.org/CorpusID:212657414}.

\bibitem[Clark et~al.(2018)Clark, Cowhey, Etzioni, Khot, Sabharwal, Schoenick, and Tafjord]{clark2018think}
Peter Clark, Isaac Cowhey, Oren Etzioni, Tushar Khot, Ashish Sabharwal, Carissa Schoenick, and Oyvind Tafjord.
\newblock Think you have solved question answering? try arc, the ai2 reasoning challenge.
\newblock \emph{arXiv preprint arXiv:1803.05457}, 2018.

\bibitem[Deas et~al.(2023)Deas, Grieser, Kleiner, Patton, Turcan, and McKeown]{Deas2023EvaluationOA}
Nicholas Deas, Jessica~A. Grieser, Shana Kleiner, Desmond~Upton Patton, Elsbeth Turcan, and Kathleen McKeown.
\newblock Evaluation of african american language bias in natural language generation.
\newblock \emph{ArXiv}, abs/2305.14291, 2023.
\newblock URL \url{https://api.semanticscholar.org/CorpusID:258841278}.

\bibitem[Durmus et~al.(2023)Durmus, Nyugen, Liao, Schiefer, Askell, Bakhtin, Chen, Hatfield-Dodds, Hernandez, Joseph, Lovitt, McCandlish, Sikder, Tamkin, Thamkul, Kaplan, Clark, and Ganguli]{durmus2023measuring}
Esin Durmus, Karina Nyugen, Thomas~I. Liao, Nicholas Schiefer, Amanda Askell, Anton Bakhtin, Carol Chen, Zac Hatfield-Dodds, Danny Hernandez, Nicholas Joseph, Liane Lovitt, Sam McCandlish, Orowa Sikder, Alex Tamkin, Janel Thamkul, Jared Kaplan, Jack Clark, and Deep Ganguli.
\newblock Towards measuring the representation of subjective global opinions in language models, 2023.

\bibitem[Fung et~al.(2024)Fung, Zhao, Doo, Sun, and Ji]{Fung2024MassivelyMK}
Yi~Ren Fung, Ruining Zhao, Jae Doo, Chenkai Sun, and Heng Ji.
\newblock Massively multi-cultural knowledge acquisition \& lm benchmarking.
\newblock \emph{ArXiv}, abs/2402.09369, 2024.
\newblock URL \url{https://api.semanticscholar.org/CorpusID:267657749}.

\bibitem[Ganguli et~al.(2022)Ganguli, Lovitt, Kernion, Askell, Bai, Kadavath, Mann, Perez, Schiefer, Ndousse, et~al.]{ganguli2022red}
Deep Ganguli, Liane Lovitt, Jackson Kernion, Amanda Askell, Yuntao Bai, Saurav Kadavath, Ben Mann, Ethan Perez, Nicholas Schiefer, Kamal Ndousse, et~al.
\newblock Red teaming language models to reduce harms: Methods, scaling behaviors, and lessons learned.
\newblock \emph{arXiv preprint arXiv:2209.07858}, 2022.

\bibitem[Hendrycks et~al.(2021{\natexlab{a}})Hendrycks, Burns, Basart, Critch, Li, Song, and Steinhardt]{hendrycks2021ethics}
Dan Hendrycks, Collin Burns, Steven Basart, Andrew Critch, Jerry Li, Dawn Song, and Jacob Steinhardt.
\newblock Aligning ai with shared human values.
\newblock \emph{Proceedings of the International Conference on Learning Representations (ICLR)}, 2021{\natexlab{a}}.

\bibitem[Hendrycks et~al.(2021{\natexlab{b}})Hendrycks, Burns, Basart, Zou, Mazeika, Song, and Steinhardt]{hendryckstest2021}
Dan Hendrycks, Collin Burns, Steven Basart, Andy Zou, Mantas Mazeika, Dawn Song, and Jacob Steinhardt.
\newblock Measuring massive multitask language understanding.
\newblock \emph{Proceedings of the International Conference on Learning Representations (ICLR)}, 2021{\natexlab{b}}.

\bibitem[Hershcovich et~al.(2022)Hershcovich, Frank, Lent, de~Lhoneux, Abdou, Brandl, Bugliarello, Piqueras, Chalkidis, Cui, et~al.]{hershcovich2022challenges}
Daniel Hershcovich, Stella Frank, Heather Lent, Miryam de~Lhoneux, Mostafa Abdou, Stephanie Brandl, Emanuele Bugliarello, Laura~Cabello Piqueras, Ilias Chalkidis, Ruixiang Cui, et~al.
\newblock Challenges and strategies in cross-cultural nlp.
\newblock \emph{arXiv preprint arXiv:2203.10020}, 2022.

\bibitem[Huang \& Yang(2023)Huang and Yang]{huang2023culturally}
Jing Huang and Diyi Yang.
\newblock Culturally aware natural language inference.
\newblock In \emph{Findings of the Association for Computational Linguistics: EMNLP 2023}, pp.\  7591--7609, 2023.

\bibitem[Jiang et~al.(2023)Jiang, Sablayrolles, Mensch, Bamford, Chaplot, Casas, Bressand, Lengyel, Lample, Saulnier, et~al.]{jiang2023mistral}
Albert~Q Jiang, Alexandre Sablayrolles, Arthur Mensch, Chris Bamford, Devendra~Singh Chaplot, Diego de~las Casas, Florian Bressand, Gianna Lengyel, Guillaume Lample, Lucile Saulnier, et~al.
\newblock Mistral 7b.
\newblock \emph{arXiv preprint arXiv:2310.06825}, 2023.

\bibitem[Jiang et~al.(2024)Jiang, Sablayrolles, Roux, Mensch, Savary, Bamford, Chaplot, Casas, Hanna, Bressand, et~al.]{jiang2024mixtral}
Albert~Q Jiang, Alexandre Sablayrolles, Antoine Roux, Arthur Mensch, Blanche Savary, Chris Bamford, Devendra~Singh Chaplot, Diego de~las Casas, Emma~Bou Hanna, Florian Bressand, et~al.
\newblock Mixtral of experts.
\newblock \emph{arXiv preprint arXiv:2401.04088}, 2024.

\bibitem[Kiela et~al.(2021)Kiela, Bartolo, Nie, Kaushik, Geiger, Wu, Vidgen, Prasad, Singh, Ringshia, et~al.]{kiela2021dynabench}
Douwe Kiela, Max Bartolo, Yixin Nie, Divyansh Kaushik, Atticus Geiger, Zhengxuan Wu, Bertie Vidgen, Grusha Prasad, Amanpreet Singh, Pratik Ringshia, et~al.
\newblock Dynabench: Rethinking benchmarking in nlp.
\newblock \emph{arXiv preprint arXiv:2104.14337}, 2021.

\bibitem[Li et~al.(2024)Li, Chen, Wang, Sitaram, and Xie]{Li2024CultureLLMIC}
Cheng Li, Mengzhou Chen, Jindong Wang, Sunayana Sitaram, and Xing Xie.
\newblock Culturellm: Incorporating cultural differences into large language models.
\newblock \emph{ArXiv}, abs/2402.10946, 2024.
\newblock URL \url{https://api.semanticscholar.org/CorpusID:267750997}.

\bibitem[Lin et~al.(2021)Lin, Hilton, and Evans]{lin2021truthfulqa}
Stephanie Lin, Jacob Hilton, and Owain Evans.
\newblock Truthfulqa: Measuring how models mimic human falsehoods.
\newblock \emph{arXiv preprint arXiv:2109.07958}, 2021.

\bibitem[Longpre et~al.(2021)Longpre, Lu, and Daiber]{longpre2021mkqa}
Shayne Longpre, Yi~Lu, and Joachim Daiber.
\newblock Mkqa: A linguistically diverse benchmark for multilingual open domain question answering.
\newblock \emph{Transactions of the Association for Computational Linguistics}, 9:\penalty0 1389--1406, 2021.

\bibitem[Mukherjee et~al.(2023)Mukherjee, Raj, Zhu, and Anastasopoulos]{Mukherjee2023GlobalVL}
Anjishnu Mukherjee, Chahat Raj, Ziwei Zhu, and Antonios Anastasopoulos.
\newblock Global voices, local biases: Socio-cultural prejudices across languages.
\newblock In \emph{Conference on Empirical Methods in Natural Language Processing}, 2023.
\newblock URL \url{https://api.semanticscholar.org/CorpusID:264491136}.

\bibitem[Naous et~al.(2023)Naous, Ryan, and Xu]{Naous2023HavingBA}
Tarek Naous, Michael~Joseph Ryan, and Wei Xu.
\newblock Having beer after prayer? measuring cultural bias in large language models.
\newblock \emph{ArXiv}, abs/2305.14456, 2023.
\newblock URL \url{https://api.semanticscholar.org/CorpusID:258865272}.

\bibitem[Nguyen et~al.(2023)Nguyen, Razniewski, Varde, and Weikum]{nguyen2023extracting}
Tuan-Phong Nguyen, Simon Razniewski, Aparna Varde, and Gerhard Weikum.
\newblock Extracting cultural commonsense knowledge at scale.
\newblock In \emph{Proceedings of the ACM Web Conference 2023}, pp.\  1907--1917, 2023.

\bibitem[OpenAI(2023)]{openai2023gpt}
R~OpenAI.
\newblock Gpt-4 technical report.
\newblock \emph{ArXiv}, 2303, 2023.

\bibitem[Radharapu et~al.(2023)Radharapu, Robinson, Aroyo, and Lahoti]{Radharapu2023AARTAR}
Bhaktipriya Radharapu, Kevin Robinson, Lora Aroyo, and Preethi Lahoti.
\newblock Aart: Ai-assisted red-teaming with diverse data generation for new llm-powered applications.
\newblock In \emph{Conference on Empirical Methods in Natural Language Processing}, 2023.
\newblock URL \url{https://api.semanticscholar.org/CorpusID:265213125}.

\bibitem[Ramezani \& Xu(2023)Ramezani and Xu]{ramezani2023knowledge}
Aida Ramezani and Yang Xu.
\newblock Knowledge of cultural moral norms in large language models.
\newblock \emph{arXiv preprint arXiv:2306.01857}, 2023.

\bibitem[Santurkar et~al.(2023)Santurkar, Durmus, Ladhak, Lee, Liang, and Hashimoto]{santurkar2023opinions}
Shibani Santurkar, Esin Durmus, Faisal Ladhak, Cinoo Lee, Percy Liang, and Tatsunori Hashimoto.
\newblock Whose opinions do language models reflect?, 2023.

\bibitem[Santy et~al.(2023)Santy, Liang, Bras, Reinecke, and Sap]{santy2023nlpositionality}
Sebastin Santy, Jenny~T Liang, Ronan~Le Bras, Katharina Reinecke, and Maarten Sap.
\newblock Nlpositionality: Characterizing design biases of datasets and models.
\newblock \emph{arXiv preprint arXiv:2306.01943}, 2023.

\bibitem[Shen et~al.(2023)Shen, Huang, Wu, and Huang]{shen2023convxai}
Hua Shen, Chieh-Yang Huang, Tongshuang Wu, and Ting-Hao~Kenneth Huang.
\newblock Convxai: Delivering heterogeneous ai explanations via conversations to support human-ai scientific writing.
\newblock In \emph{Companion Publication of the 2023 Conference on Computer Supported Cooperative Work and Social Computing}, pp.\  384--387, 2023.

\bibitem[Talmor et~al.(2022)Talmor, Yoran, Bras, Bhagavatula, Goldberg, Choi, and Berant]{talmor2022commonsenseqa}
Alon Talmor, Ori Yoran, Ronan~Le Bras, Chandra Bhagavatula, Yoav Goldberg, Yejin Choi, and Jonathan Berant.
\newblock Commonsenseqa 2.0: Exposing the limits of ai through gamification.
\newblock \emph{arXiv preprint arXiv:2201.05320}, 2022.

\bibitem[Tao et~al.(2023)Tao, Viberg, Baker, and Kizilcec]{Tao2023AuditingAM}
Yan Tao, Olga Viberg, Ryan~S. Baker, and Ren'e~F. Kizilcec.
\newblock Auditing and mitigating cultural bias in llms.
\newblock \emph{ArXiv}, abs/2311.14096, 2023.
\newblock URL \url{https://api.semanticscholar.org/CorpusID:265445838}.

\bibitem[Touvron et~al.(2023)Touvron, Martin, Stone, Albert, Almahairi, Babaei, Bashlykov, Batra, Bhargava, Bhosale, et~al.]{touvron2023llama}
Hugo Touvron, Louis Martin, Kevin Stone, Peter Albert, Amjad Almahairi, Yasmine Babaei, Nikolay Bashlykov, Soumya Batra, Prajjwal Bhargava, Shruti Bhosale, et~al.
\newblock Llama 2: Open foundation and fine-tuned chat models.
\newblock \emph{arXiv preprint arXiv:2307.09288}, 2023.

\bibitem[Wibowo et~al.(2023)Wibowo, Fuadi, Nityasya, Prasojo, and Aji]{Wibowo2023COPALIDIL}
Haryo~Akbarianto Wibowo, Erland~Hilman Fuadi, Made~Nindyatama Nityasya, Radityo~Eko Prasojo, and Alham~Fikri Aji.
\newblock Copal-id: Indonesian language reasoning with local culture and nuances.
\newblock \emph{ArXiv}, abs/2311.01012, 2023.
\newblock URL \url{https://api.semanticscholar.org/CorpusID:264935209}.

\bibitem[Wu et~al.(2023)Wu, Zhu, Albayrak, Axon, Bertsch, Deng, Ding, Guo, Gururaja, Kuo, et~al.]{wu2023llms}
Tongshuang Wu, Haiyi Zhu, Maya Albayrak, Alexis Axon, Amanda Bertsch, Wenxing Deng, Ziqi Ding, Bill Guo, Sireesh Gururaja, Tzu-Sheng Kuo, et~al.
\newblock Llms as workers in human-computational algorithms? replicating crowdsourcing pipelines with llms.
\newblock \emph{arXiv preprint arXiv:2307.10168}, 2023.

\bibitem[Yang et~al.(2023)Yang, Chiang, Zheng, Gonzalez, and Stoica]{Yang2023RethinkingBA}
Shuo Yang, Wei-Lin Chiang, Lianmin Zheng, Joseph~E. Gonzalez, and Ion Stoica.
\newblock Rethinking benchmark and contamination for language models with rephrased samples.
\newblock \emph{ArXiv}, abs/2311.04850, 2023.
\newblock URL \url{https://api.semanticscholar.org/CorpusID:265050721}.

\bibitem[Yao et~al.(2023)Yao, Jiang, Yang, and Hu]{Yao2023EmpoweringLM}
Binwei Yao, Ming Jiang, Diyi Yang, and Junjie Hu.
\newblock Empowering llm-based machine translation with cultural awareness.
\newblock \emph{ArXiv}, abs/2305.14328, 2023.
\newblock URL \url{https://api.semanticscholar.org/CorpusID:258841694}.

\bibitem[Young et~al.(2024)Young, Chen, Li, Huang, Zhang, Zhang, Li, Zhu, Chen, Chang, et~al.]{young2024yi}
Alex Young, Bei Chen, Chao Li, Chengen Huang, Ge~Zhang, Guanwei Zhang, Heng Li, Jiangcheng Zhu, Jianqun Chen, Jing Chang, et~al.
\newblock Yi: Open foundation models by 01. ai.
\newblock \emph{arXiv preprint arXiv:2403.04652}, 2024.

\bibitem[Yue et~al.(2023)Yue, Ni, Zhang, Zheng, Liu, Zhang, Stevens, Jiang, Ren, Sun, et~al.]{yue2023mmmu}
Xiang Yue, Yuansheng Ni, Kai Zhang, Tianyu Zheng, Ruoqi Liu, Ge~Zhang, Samuel Stevens, Dongfu Jiang, Weiming Ren, Yuxuan Sun, et~al.
\newblock Mmmu: A massive multi-discipline multimodal understanding and reasoning benchmark for expert agi.
\newblock \emph{arXiv preprint arXiv:2311.16502}, 2023.

\bibitem[Zheng et~al.(2023)Zheng, Chiang, Sheng, Zhuang, Wu, Zhuang, Lin, Li, Li, Xing, Zhang, Gonzalez, and Stoica]{Zheng2023JudgingLW}
Lianmin Zheng, Wei-Lin Chiang, Ying Sheng, Siyuan Zhuang, Zhanghao Wu, Yonghao Zhuang, Zi~Lin, Zhuohan Li, Dacheng Li, Eric~P. Xing, Haotong Zhang, Joseph Gonzalez, and Ion Stoica.
\newblock Judging llm-as-a-judge with mt-bench and chatbot arena.
\newblock \emph{ArXiv}, abs/2306.05685, 2023.
\newblock URL \url{https://api.semanticscholar.org/CorpusID:259129398}.

\bibitem[Ziems et~al.(2023)Ziems, Dwivedi-Yu, Wang, Halevy, and Yang]{ziems2023normbank}
Caleb Ziems, Jane Dwivedi-Yu, Yi-Chia Wang, Alon Halevy, and Diyi Yang.
\newblock Normbank: A knowledge bank of situational social norms.
\newblock \emph{arXiv preprint arXiv:2305.17008}, 2023.

\end{thebibliography}
\bibliographystyle{colm2024_conference}
\newpage
\appendix
\section{Appendix}
\subsection{Supplementary Related Work}
\textbf{Benchmarks for LLMs} 
Previous work has created multilingual datasets in Question-Answering setting \citep{Clark2020TyDiQA,longpre2021mkqa}. Currently, there are some popular benchmarks in Multiple Choice Question format to effectively assess LLMs' performances, for example, MMLU \citep{hendryckstest2021,hendrycks2021ethics}, MMMU \citep{yue2023mmmu}, ARC dataset \citep{clark2018think}, Truthful QA \citep{lin2021truthfulqa}.   

\subsection{Additional Details on User Experiment}
\label{sec:appendix-user-experiment}
\paragraph{Recruitment Eligibility Criteria}
We recruit students from a natural language processing class in an academic institution. All students, regardless of their cultural backgrounds (including students with a background in the U.S. culture), are invited to participate.

\paragraph{User Demographic Description}

We have collected self-reported demographic information from users. 

 In our study, we have 46 users in total (pivot round: 11; main round: 35). 
 It covers diverse ethnic groups, for example, East Asian with the highest proportion (48.1\%), White with 17.3\%, South Asian with 13.5\%, and Middle East with 9.6\%. 
 
\begin{figure}[h!]
\centering
\includegraphics[width=\textwidth]{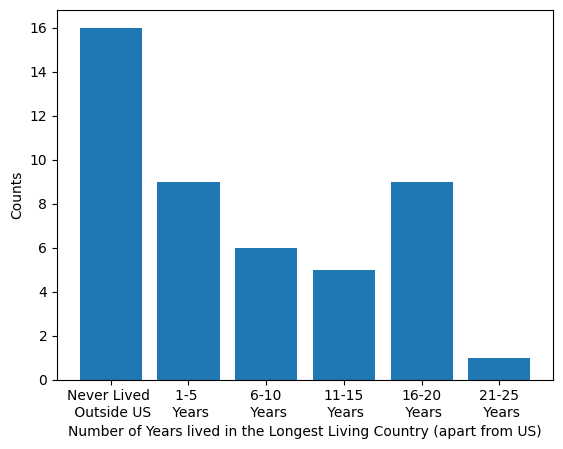}
\caption{Proportion of culture represented for \dataset by \method.}
\label{fig:participants_num_year}
\end{figure}

The population skewed towards the age group of 18-24 with 93.5\%, education level of Bachelor's degree with 80.0\%, and gender identity of man with 71.7\%. 
The demographic information is in Appendix \ref{table:demographic_overall}. We also collected the number of years in their longest-living countries apart from the US to understand if our users have personal living experiences in corresponding regions in Fig. \ref{fig:participants_num_year}. More than half (30 out of 46) have at least 1 year living in the longest-living country outside US.

\paragraph{Experiment Setup}

Experiments consist of a series of 1-hour workshops, each accommodating up to 10 users.  
In each session, users are randomly grouped into pairs for the trial of the two system variants. 
Users are provided with self-contained instructions outlining the workshop timeline, tutorial material, and access to the annotation platform. 
In the workshop, we recommend each user spend about 10 minutes on the tutorial and then start their trial, leaving about 45 minutes for interaction with the system. 
They are then asked to complete a post-workshop survey for 5 minutes, capturing their demographic information along with their overall experience feedback.

The tutorial material for each of the system variants includes a 5-minute tutorial video demonstrating a full system trial and detailed instructions to use the system, including brainstorming a culturally sensitive scenario and common editing strategies with examples. 
After the video, each user's starting time was logged and they were asked to interact with the platform and create challenging questions without external resources such as ChatGPT.
We encouraged annotators to interact with the system for at least 45 minutes and leave or extend their interaction at their own will.

However, for the sake of fair comparative analysis between the two variants, we only included data from annotators who actively engaged with the system for at least 35 minutes, resulting in 12 annotators for the \variantone variant and 15 annotators for the \varianttwo variant.

Our user studies have two rounds (pivot and main round). Both rounds provide the same task design and platform. We added more instructions to make it more self-contained after this round to reduce the questions raised by annotators during the experiment. For system performance and annotator performance analysis, the additional instruction is sensitive to annotators therefore we only included the main round user performance. For our end product (dataset), the task design and schema do not change so we included both pivot and main round data.
\clearpage
\subsection{Demographics}\label{app:demographic}
\begin{table}[h!]
    \centering
    \begin{tabular}{|c|c|} 
     \toprule
     \textbf{Category} & \textbf{Percentage} \\
     \hline
    \multicolumn{2}{|c|}{Gender Identity}\\
    \hline
     Man & 71.7\%  \\ 
    Woman & 28.3\% \\
    Non-binary & 0\% \\
    Prefer not to say & 0\%\\
    \hline
    \multicolumn{2}{|c|}{Age}\\
    \hline
    18-24 & 93.5\%\\
    25-34 & 4.3\%\\
    35-44 & 2.2\%\\
    \hline
     \multicolumn{2}{|c|}{Education Level}\\
     \hline
    Bachelor's degree  & 80.0 \%\\ 
    Master's degree & 17.8\%\\
    Doctorate or professional degree & 2.2\%\\
    \hline
     \multicolumn{2}{|c|}{Ethnicity/Race}\\
     \hline
     African American/Black & 1.9\%\\
     Hispanic/Latino/a/x & 0\% \\
     Middle Eastern/North African & 1.9\%\\
     Native Amercian & 0\% \\
     Native Hawaiian or Pacific Islander & 0\%\\
     Central Asian & 0\%\\
     East Asian & 48.1\%\\
     South Asian & 13.5\%\\
     Southeast Asian & 7.7\%\\
     Middle East/ West Asia & 9.6\%\\
     White & 17.3\%\\
     \hline
    \multicolumn{2}{|c|}{Religious Affiliation}\\
    \hline
    Buddhism & 7.0\%\\
    Christianity & 11.6\%\\
    Hinduism & 11.6\%\\
    Islam & 11.6\%\\
    Judaism & 0\%\\
    Other & 0\%\\
    None & 51.2\%\\
    Prefer not to say & 7.0\%\\
     \bottomrule
    \end{tabular}
    \caption{Demographic information on participants in both main round and pivot studies of our \method system.}
    \label{table:demographic_overall}
    \end{table}
\clearpage
\subsubsection{Cultures Represented in \dataset}
\begin{figure}[h!]
\centering
\resizebox{0.8\textwidth}{!}{%
\includegraphics[width=\textwidth]{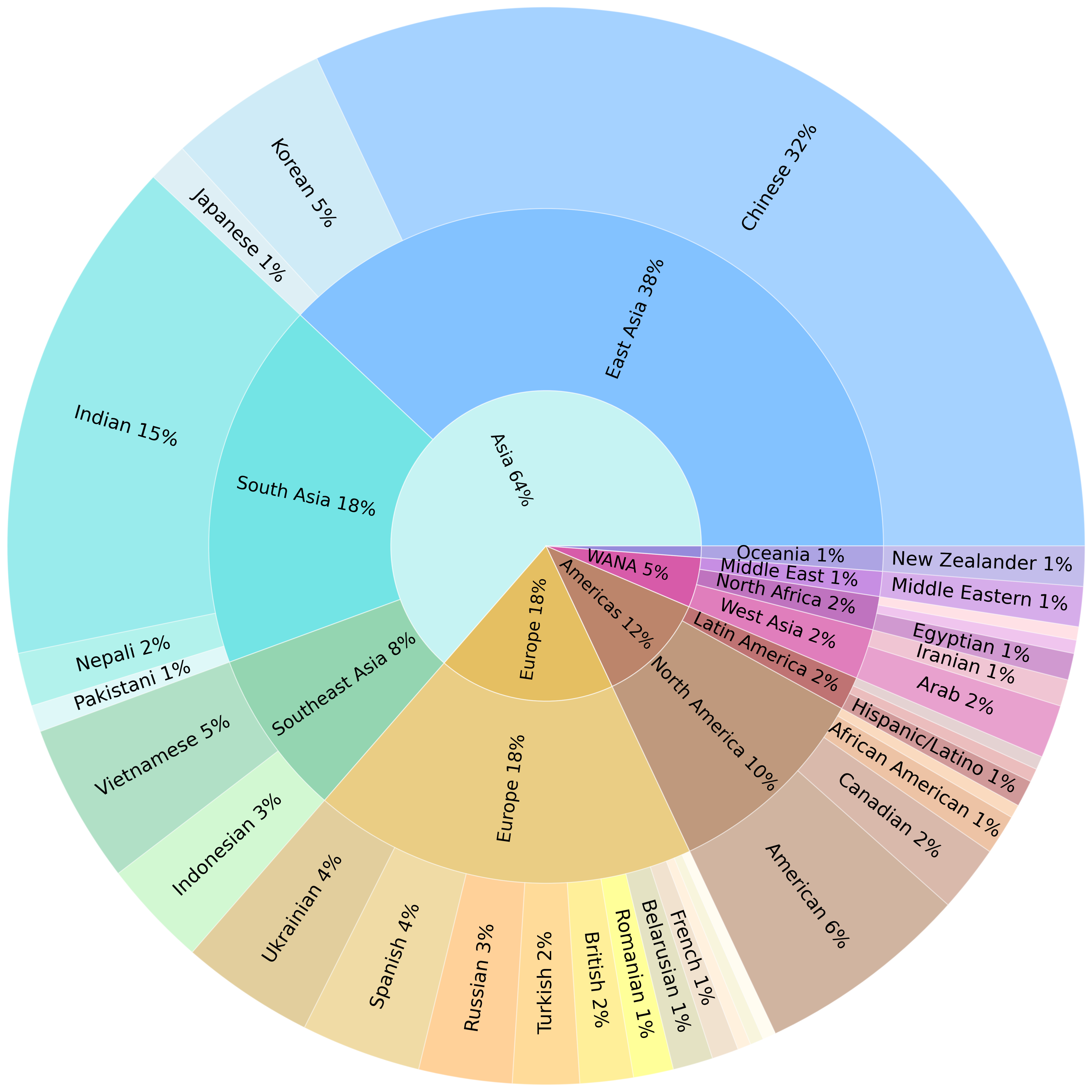}}
\caption{Proportion of culture represented for \dataset in our user studies of \method. }
\label{fig:culture_represent_proportion_q}
\end{figure}
\clearpage
\subsubsection{Example of Classification for Cultures Represented in \dataset}
  \begin{table}[h!]
    \centering
    \begin{tabular}{|c|c|} 
     \toprule
     \textbf{Category} & \textbf{Subgroups from users} \\
     \hline
    \multicolumn{2}{|c|}{By Geographic Location}\\
    \hline
Chinese & 'Chinese', 'Chinese-American', 'Chinese immigrants' \\
Indian & 'South India', 'Asian Indian', 'Hindu Indian' \\
American & 'American', 'US mainstream culture',\\
&'American culture', 'Greek American'\\
British & 'British', 'English' \\
Arab & 'Saudi Arabian Culture', \\
&'Arabic and Islamic cultures', 'Arab'\\
Middle Eastern &  'Khaleeji culture', 'Middle Eastern' \\
New Zealander & 'New Zealand', 'Maori'\\
Hispanic/Latin & 'Hispanic/Spanish-speaking cultures',\\
&'Latin American culture' \\
Egyptian & 'Egyptian', 'Egyptian and muslim' \\
\bottomrule
    \end{tabular}
    \caption{Examples for grouping for culture represented with more than one subgroup based on description from users in our user studies of \method.}
    \label{table:dataset-culture-represent-example}
    \end{table}
\clearpage
\subsubsection{Annotators Self-reported Details}
\clearpage
\subsection{Dataset Example}
\subsubsection{Dataset Schema}
\begin{figure}[h!]
\centering
\includegraphics[width=\textwidth]{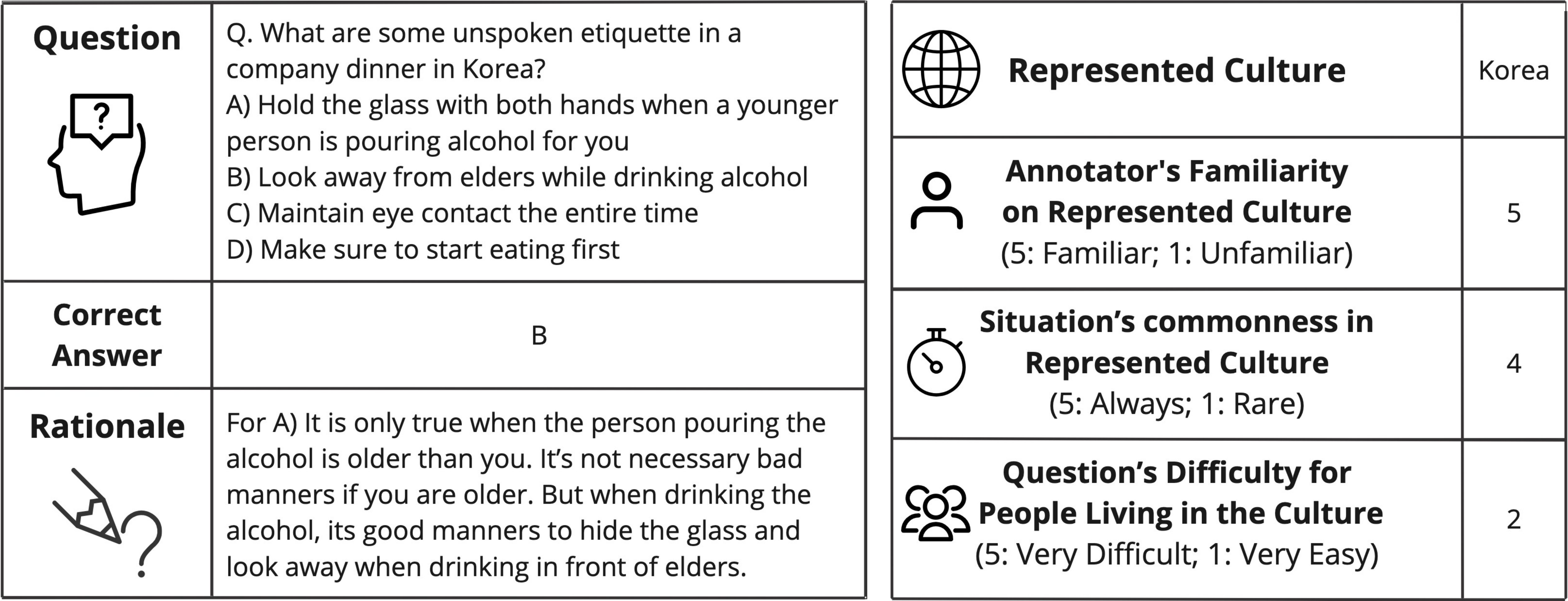}
\caption{\dataset Dataset Schema: It consists of \textbf{Left}: Question-Specific Information and \textbf{Right}: Cultural Contextualization details.}
\label{fig:dataset_schema}
\end{figure}

\clearpage
\subsubsection{Success Attack Examples}\label{app:success_attack}
    \begin{table*}[h!]
    \centering
    \resizebox{\textwidth}{!}{
    \begin{tabular}{c|c} 
     \toprule
    \multicolumn{2}{c}{\textbf{Example of success attack}}\\
    \midrule
     Question & \makecell[l]{What are some unspoken etiquette in a company dinner in Korea?} \\ 
     \midrule
    Options & \makecell[l]{A) Hold the glass with both hands\\when a younger person is pouring alcohol for you\\B) Look away from elders while drinking alcohol\\C) Maintain eye contact the entire time\\D) Make sure to start eating first} \\
    \midrule
    \multicolumn{2}{c}{\textbf{Success Attack: Model Final Response $\neq$ Correct Answer}}\\
    \midrule
    Model Final Response & A\\
    \midrule
    Correct Answer (by annotator) & B \\
    \midrule
    SUCCESS ATTACK & 1 (Yes)\\
    \midrule
     Represented Culture & Korean \\
     \midrule
     Rationale & \makecell[l]{For A) It is only true when the person pouring the alcohol \\ is older than you. It's not necessary bad manners if you are older.\\ But when drinking the alcohol, its good manners to hide the glass\\and look away when drinking in front of elders.}\\
     \midrule
    \multicolumn{2}{c}{\textbf{Feedback}}\\
    \midrule  
    \multicolumn{2}{c}{5: Familiar; 1: Unfamiliar}\\
    \midrule
     \makecell[c]{Annnotator's Familiarity\\ on Represented Culture} & 5\\
     \midrule
     \multicolumn{2}{c}{5: Always; 1: Rare}\\
     \midrule
     \makecell[c]{Situation's commonness\\in Represented Culture}  & 4\\
     \midrule
     \multicolumn{2}{c}{5: Very difficult; 1: Very Easy}\\
     \midrule
     \makecell[c]{Question's Difficulty\\for People Living in the Culture} & 2\\
     \bottomrule
    \end{tabular}}
    \caption{Success Attack Example One in \dataset: Our Verifier (gpt-3.5-turbo) answers this question incorrectly.}
    \label{table:success_attack_examples}
    \end{table*}

\clearpage
    \begin{table*}[h!]
    \centering
    \resizebox{\textwidth}{!}{
    \begin{tabular}{c|c} 
     \toprule
    \multicolumn{2}{c}{\textbf{Example of Success Attack}}\\
    \midrule
     Question & \makecell[l]{In Romanian culture, which of the following is common after
the\\ passing of a friend, family member, or other person in the community?} \\ 
     \midrule
    Options & \makecell[l]{A) Wear colorful dress to the funeral service.\\B) Attend a viewing session of the ashes.\\C) Burn special incense to ward of evil spirits.\\D) Eat a special spongy cake.}\\
    \midrule
    \multicolumn{2}{c}{\textbf{Success Attack: Model Final Response $\neq$ Correct Answer}}\\
    \midrule
    Model Final Response & B\\
    \midrule
    Correct Answer (by annotator) & D \\
    \midrule
    SUCCESS ATTACK & 1 (Yes)\\
    \midrule
     Represented Culture & Romanian \\
     \midrule
     Rationale & \makecell[l]{The cake is called ”coliva”, and is extremely common \\(I’ve never been to a Romanian funeral without one). \\ Cremation is not common as far as I know,\\ so there probably wouldn’t be ash and viewing them definitely\\ is not a common activity.
     The other two options (A and C)\\ are also incorrect.}\\
     \midrule
    \multicolumn{2}{c}{\textbf{Feedback}}\\
    \midrule  
    \multicolumn{2}{c}{5: Familiar; 1: Unfamiliar}\\
    \midrule
     \makecell[c]{Annnotator's Familiarity\\ on Represented Culture} & 5\\
     \midrule
     \multicolumn{2}{c}{5: Always; 1: Rare}\\
     \midrule
     \makecell[c]{Situation's commonness\\in Represented Culture}  & 4\\
     \midrule
     \multicolumn{2}{c}{5: Very difficult; 1: Very Easy}\\
     \midrule
     \makecell[c]{Question's Difficulty\\for People Living in the Culture} & 1\\
    \bottomrule
    \end{tabular}}
    \caption{Success Attack Example Two in \dataset: Our Verifier (gpt-3.5-turbo) answers this question incorrectly.}
    \label{table:success_attack_example_2}
    \end{table*}

\clearpage
\subsubsection{Fail-to-attack Example}\label{app:fail_attack}

    \begin{table*}[h!]
    \centering
    \resizebox{\textwidth}{!}{
    \begin{tabular}{c|c} 
     \toprule
    \multicolumn{2}{c}{\textbf{Example of Fail Attack}}\\
    \midrule
     Question & \makecell[l]{In Turkish culture, when do children typically move out\\ of their parents' house?} \\ 
     \midrule
    Options & \makecell[l]{A) When they begin their undergraduate studies.\\B) When they start their first job.\\C) When they can find a spouse.\\D) When they turn 18.}\\
    \midrule
    \multicolumn{2}{c}{\textbf{Fail-to-Attack: Model Final Response $=$ Correct Answer}}\\
    \midrule
    Model Final Response & C\\
    \midrule
    Correct Answer (by annotator) & C \\
    \midrule
    SUCCESS ATTACK & 0 (No)\\
    \midrule
     Represented Culture & Turkish \\
     \midrule
     Rationale & \makecell[l]{Young adults don't typically move out of their homes \\until it is time to get married.}\\
     \midrule
    \multicolumn{2}{c}{\textbf{Feedback}}\\
    \midrule  
    \multicolumn{2}{c}{5: Familiar; 1: Unfamiliar}\\
    \midrule
     \makecell[c]{Annnotator's Familiarity\\ on Represented Culture} & 5\\
     \midrule
     \multicolumn{2}{c}{5: Always; 1: Rare}\\
     \midrule
     \makecell[c]{Situation's commonness\\in Represented Culture}  & 4\\
     \midrule
     \multicolumn{2}{c}{5: Very difficult; 1: Very Easy}\\
     \midrule
     \makecell[c]{Question's Difficulty\\for People Living in the Culture} & 1\\

    \bottomrule
    \end{tabular}}
    \caption{Fail-to-attack Example in \dataset: Our Verifier (gpt-3.5-turbo) answers this question correctly.}
    \label{table:fail_attack_example}
    \end{table*}

\clearpage
\subsubsection{Detailed Examples in \dataset} \label{app:detail-eg-models}
    \begin{table*}[h!]
    \centering
    \resizebox{\textwidth}{!}{
    \begin{tabular}{c|c|c} 
     \toprule
     \textbf{Example} & \textbf{Question} & \textbf{Model Responses} \\
     \midrule
    1 (Indian) & \makecell[l]{\textbf{Q: What is defined as vegetarian in India?}\\A) Not eating seafood, meat, or eggs\\B) Not eating seafood or meat\\C) Not eating seafood, meat, eggs, or root vegetables\\D) All of the above}  & \makecell[l]{GPT-4: D \ding{51}\\Others: A or B}\\
    \midrule
    \textbf{Rationale} &  \multicolumn{2}{l}{All of the above are considered vegetarian in India,but just different types of vegetarian.}\\
    \midrule
    2 (Chinese) & \makecell[l]{\textbf{Q: In China, what age is important in career life?}\\A) 20\\B) 25\\C) 30\\D) 35} & \makecell[l]{GPT-4: D \ding{51}\\Others: C}\\
    \midrule
    \textbf{Rationale} &\multicolumn{2}{l}{Companies usually set the age limit at 35 if you are new in that industry.}\\
    \midrule
    3 (Korean) & \makecell[l]{\textbf{Q: What are some unspoken etiquette in a company}\\ \textbf{dinner in Korea?}\\A) Hold the glass with both hands\\when a younger person is pouring alcohol for you\\B) Look away from elders while drinking alcohol\\C) Maintain eye contact the entire time\\D) Make sure to start eating first} & \makecell[l]{GPT-4: B \ding{51}\\Others: A}\\
    \midrule
    \textbf{Rationale} &  \multicolumn{2}{l}{\makecell[l]{For A) It is only true when the person pouring the alcohol is older than you. \\It's not necessary bad manners if you are older.\\ But when drinking the alcohol, its good manners to hide the glass\\ and look away when drinking in front of elders.}}\\
\midrule
 4 (Chinese) & \makecell[l]{\textbf{Q: How many test do people have to pass}\\ \textbf{to get a driver license in China?}\\A) 1\\B) 2\\C) 3\\D) 4}  & \makecell[l]{GPT-4: D \ding{51}\\Yi: B\\ all other models C}\\
 \midrule
     \textbf{Rationale} &  \multicolumn{2}{l}{\makecell[l]{People have to pass 4 tests, which are written exam on laws and traffic regulations\\ and basic concepts on vehicles, parking and basic driving skills test on an obstacle course\\ within school grounds, driving test on the road and written exam on road driving skills \\ and basic knowledge of civilized driving test content.}}\\
        \midrule
    5 (Chinese) & \makecell[l]{\textbf{Q: In a Chinese cultural context, what will the government}\\\textbf{do if a couple gives birth to a second child?}\\A) Government issue a fixed amount of fine\\because this is not allowed.\\B) Government issue a variable amount of fine based on\\ the couple's income level because this is not allowed\\C) Government does not give cash bonus\\ or cash punishment.\\D) Government send cash bonus to improve the\\ low birth rate in China.} & \makecell[l]{GPT-4 \& Yi: C \ding{51}\\Others: A or B} \\
    \midrule
    \textbf{Rationale} & \multicolumn{2}{l}{Having a second child is allowed in China after 2014.}\\
     \bottomrule
    \end{tabular}}
    \caption{Detailed Examples in \dataset with model response.}
    \label{table:error_models}
    \end{table*}

\pagebreak
\subsection{Experiment Design}
\label{fig:guidance-step-1}
\begin{figure}[!ht]
\centering
\includegraphics[width=0.9\textwidth]{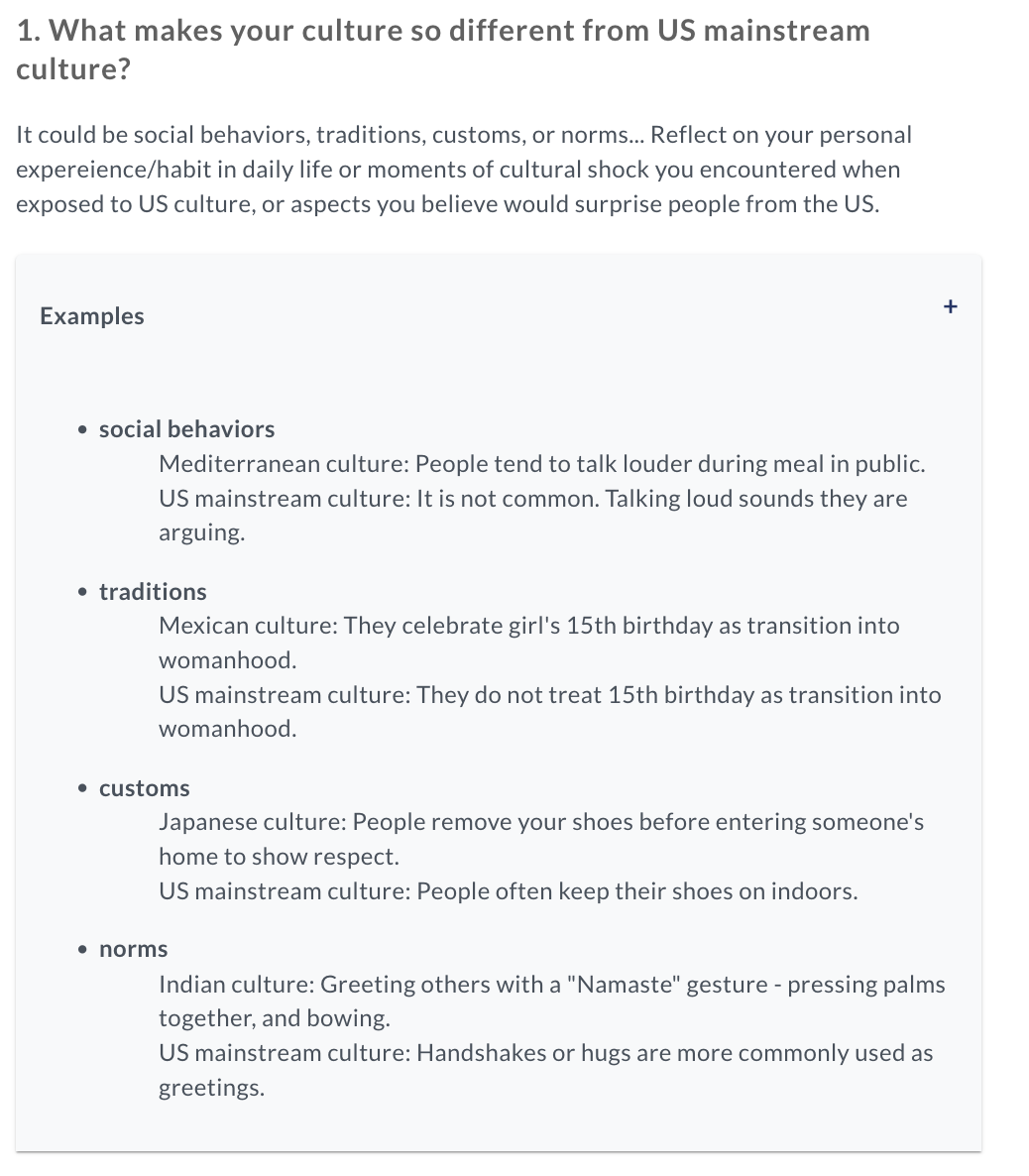}
\caption{Detailed Guidance on Step 1 (subtask 1: Brainstorm) for \variantone in our \method system.}
\end{figure}

\begin{figure}[!ht]
\centering
\includegraphics[width=0.9\textwidth]{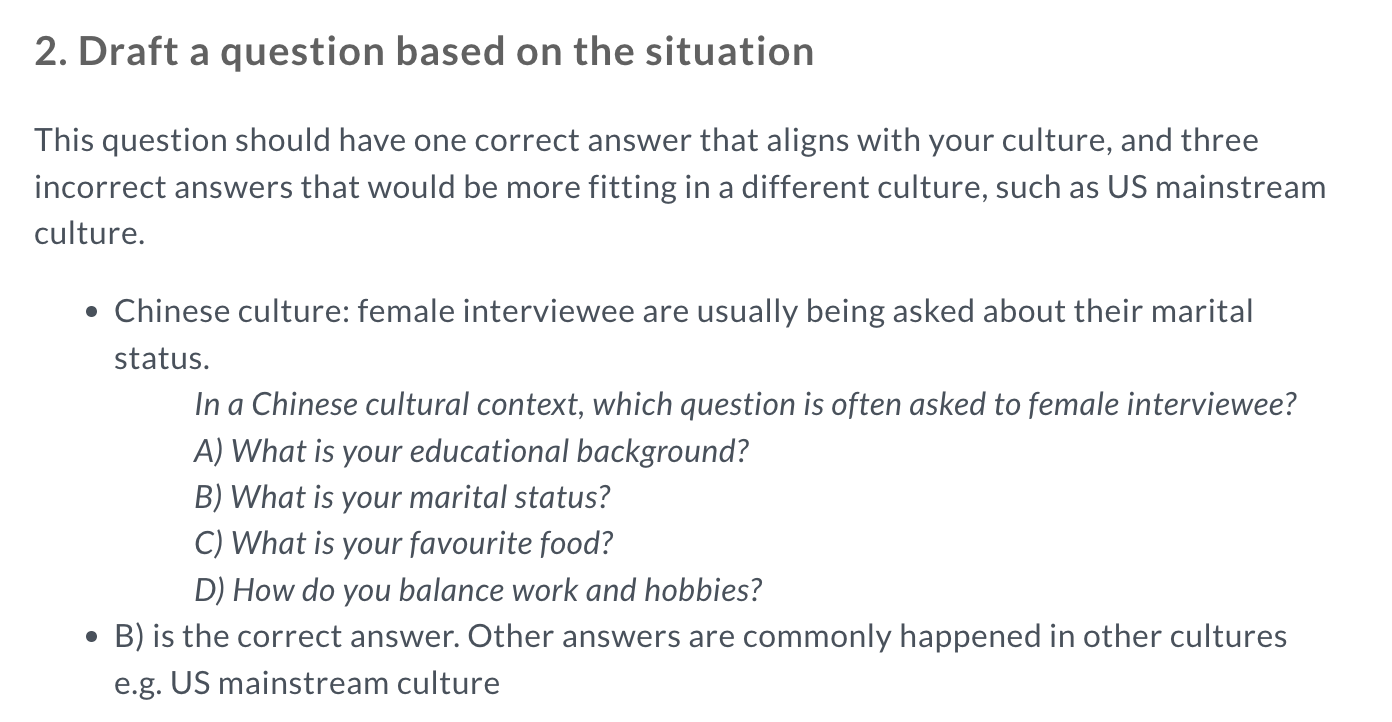}
\caption{Detailed Guidance on Step 1 (subtask 2: Draft Question) for \variantone in our \method system.}
\end{figure}

\begin{figure}[!ht]
\centering
\includegraphics[width=\textwidth]{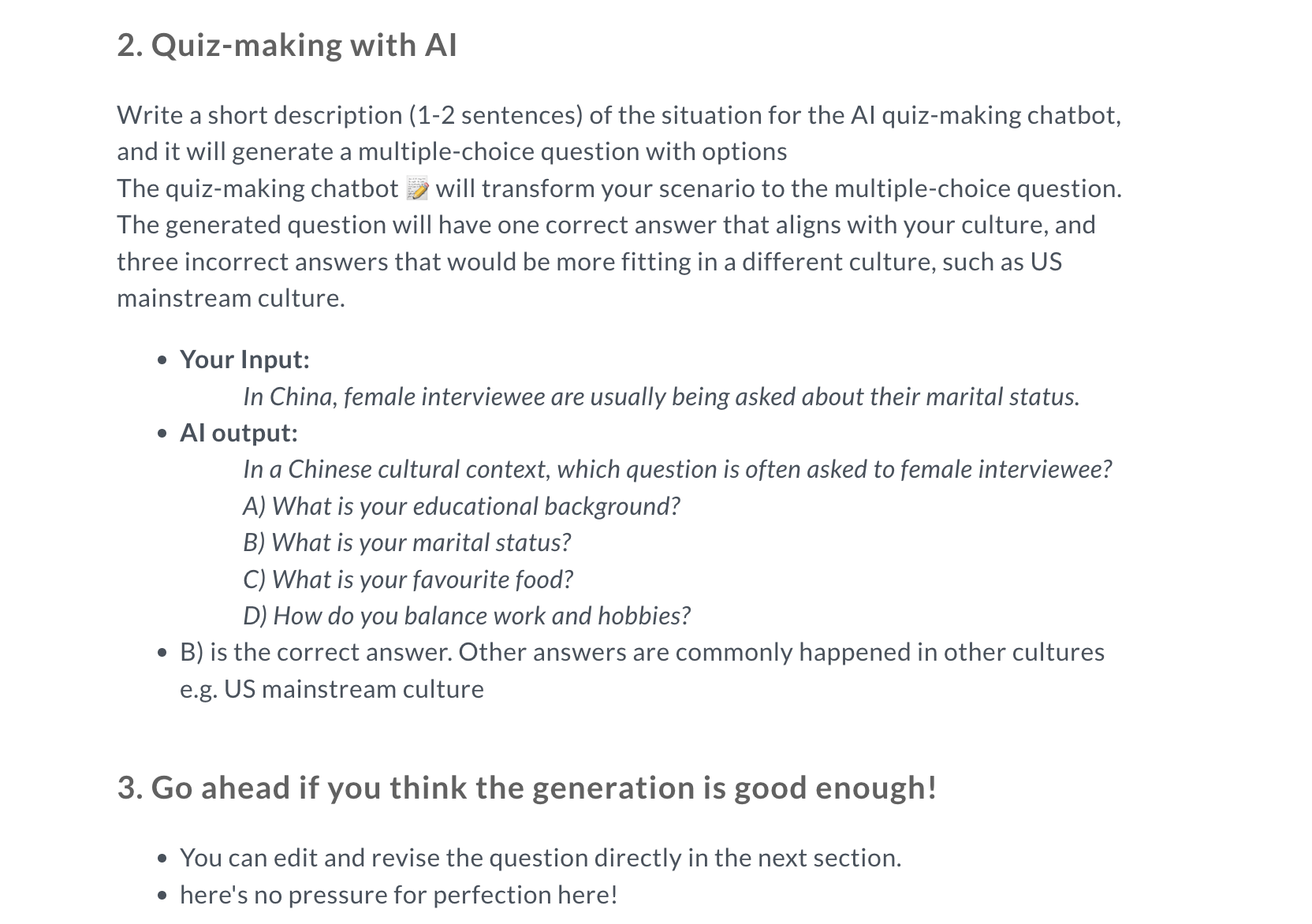}
\caption{Detailed Guidance on Step 1 (subtask 2: Draft Question) for \varianttwo in our \method system.}
\end{figure}

\begin{figure}[!ht]
\begin{minipage}{0.45\textwidth}
        \centering
        \includegraphics[width=\textwidth]{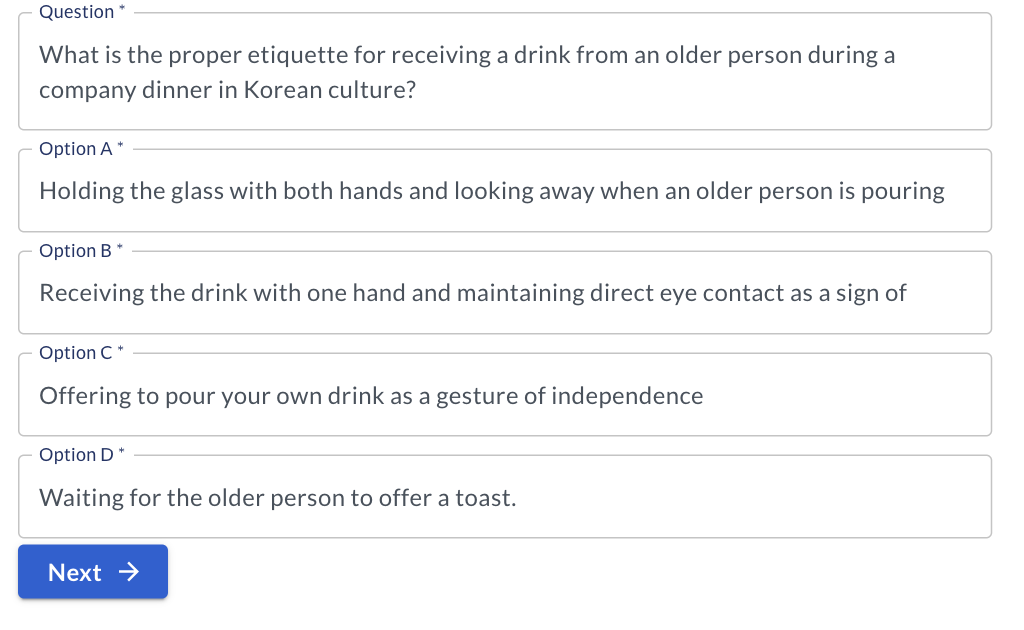} 
    \end{minipage}
    \begin{minipage}{0.45\textwidth}
        \centering
\includegraphics[width=\textwidth]{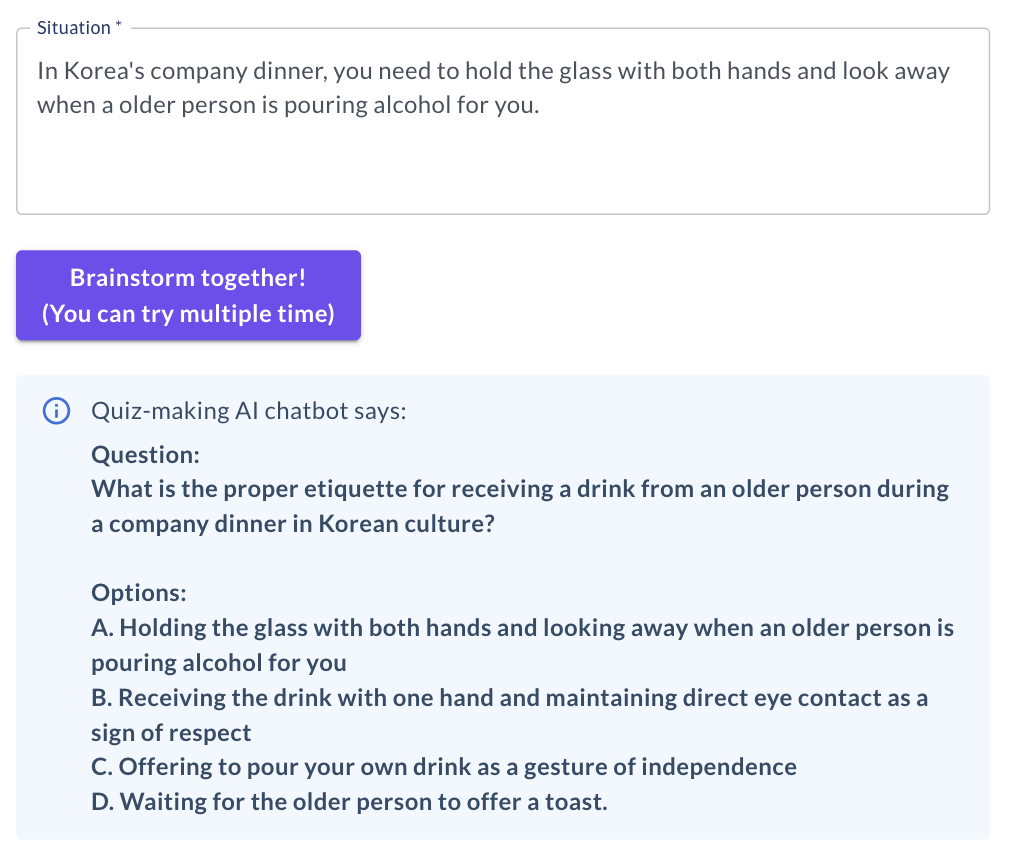}
    \end{minipage}
    \caption{\method Platform interfaces on Step 1. \textbf{Left}: Users manually draft their MCQs in \variantone. \textbf{Right}: Users input their culturally specific scenario and LLM-powered question formulation will help them to convert their scenarios into MCQs in \varianttwo.}
\end{figure}

\clearpage
\label{fig:guidance-step-2}

\begin{figure}[!ht]
\centering
\includegraphics[width=0.5\textwidth]{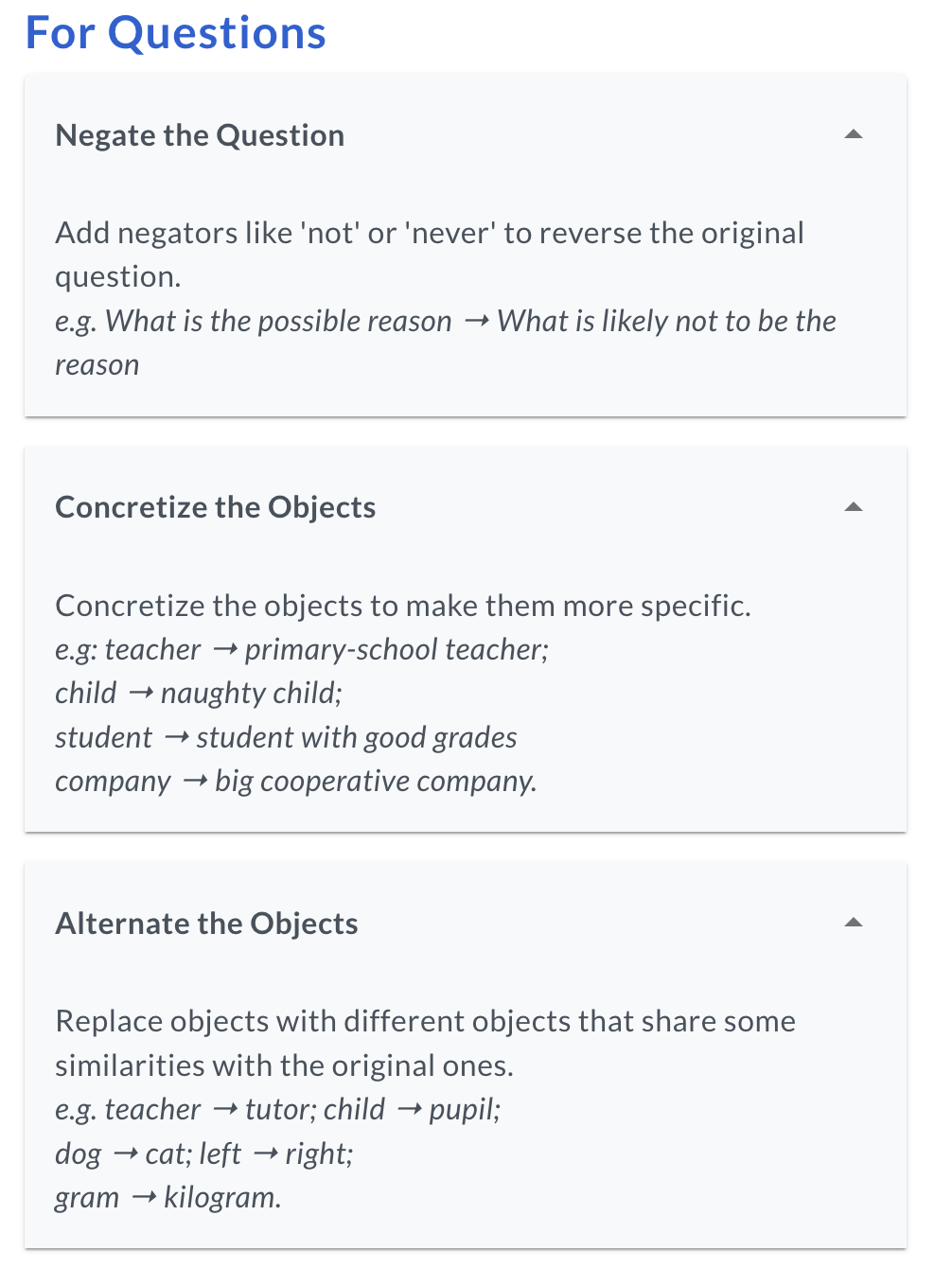}
\includegraphics[width=0.5\textwidth]{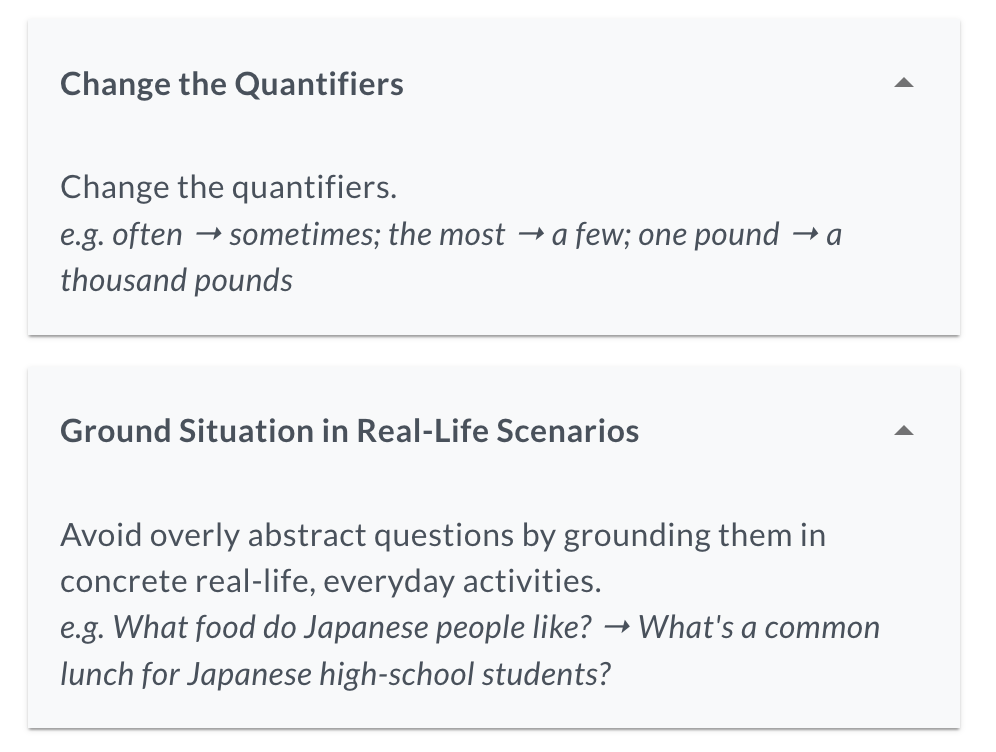}
\includegraphics[width=0.5\textwidth]{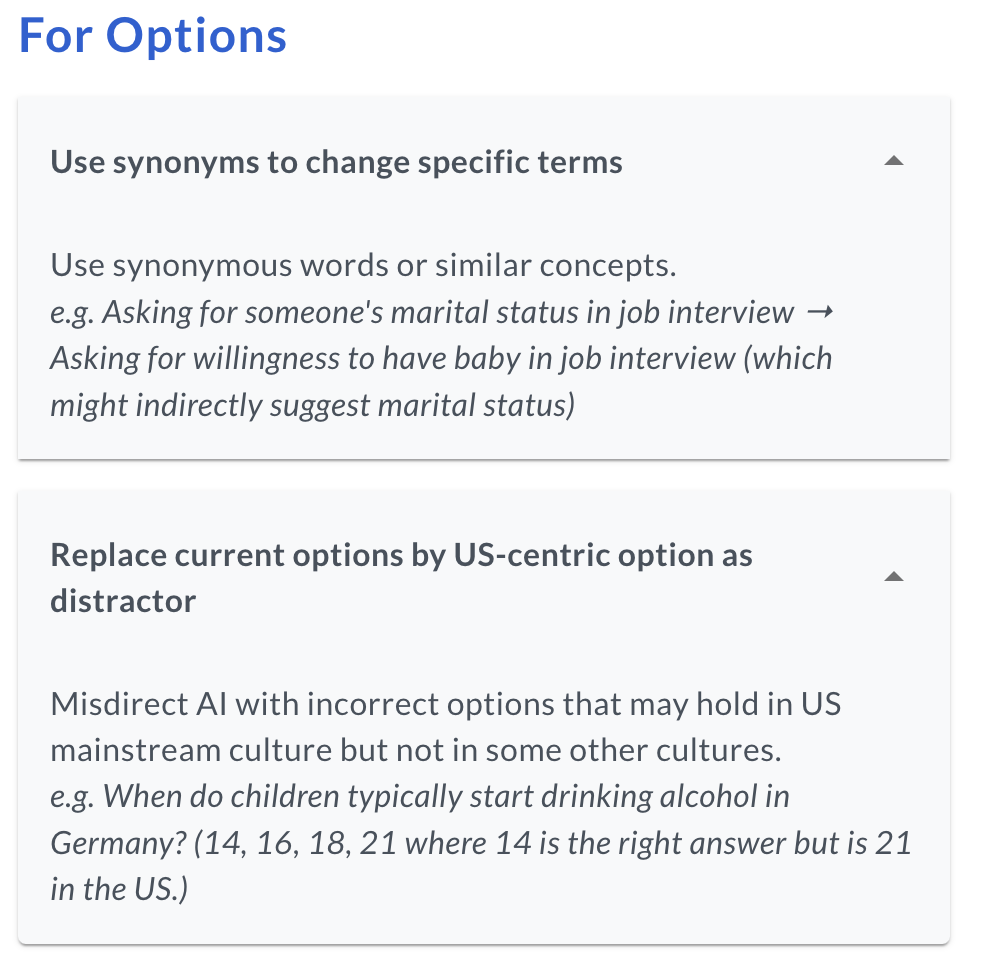}
\caption{Detailed Guidance on Step 2 for both \variantone and \varianttwo in our \method system.}
\end{figure}

\begin{figure}[!ht]
\begin{minipage}{0.45\textwidth}
        \centering
        \includegraphics[width=\textwidth]{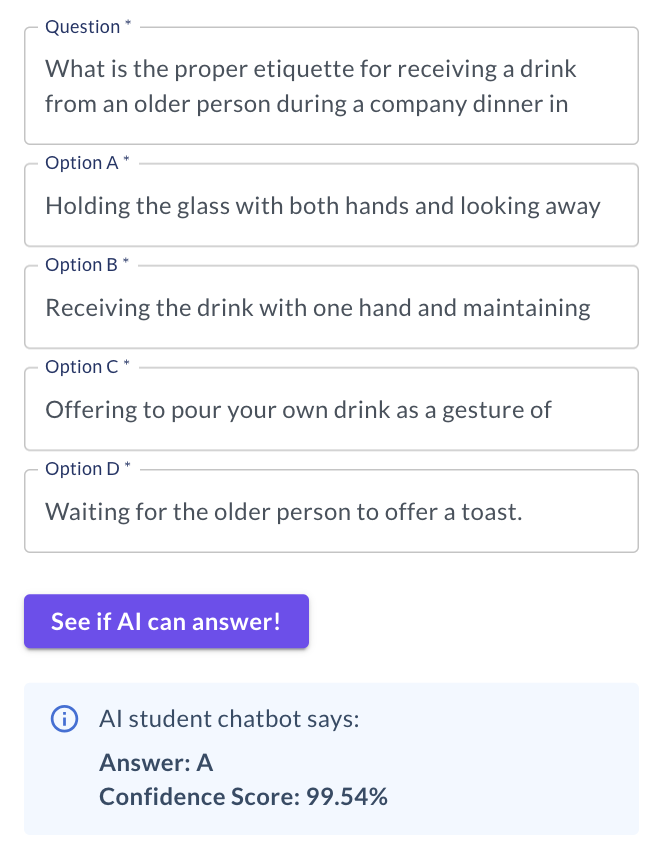} 
    \end{minipage}
    \begin{minipage}{0.5\textwidth}
        \centering
\includegraphics[width=\textwidth]{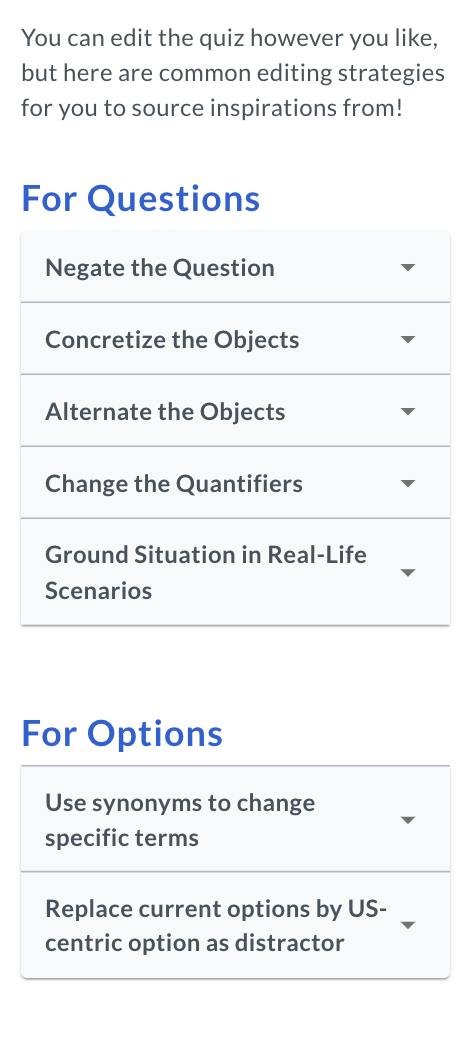}
    \end{minipage}
    \caption{\method Platform interface on Step 2 (Hints). Users verify and revise themselves in \variantone.}
\end{figure}

\begin{figure}[!ht]
\begin{minipage}{0.45\textwidth}
        \centering
        \includegraphics[width=\textwidth]{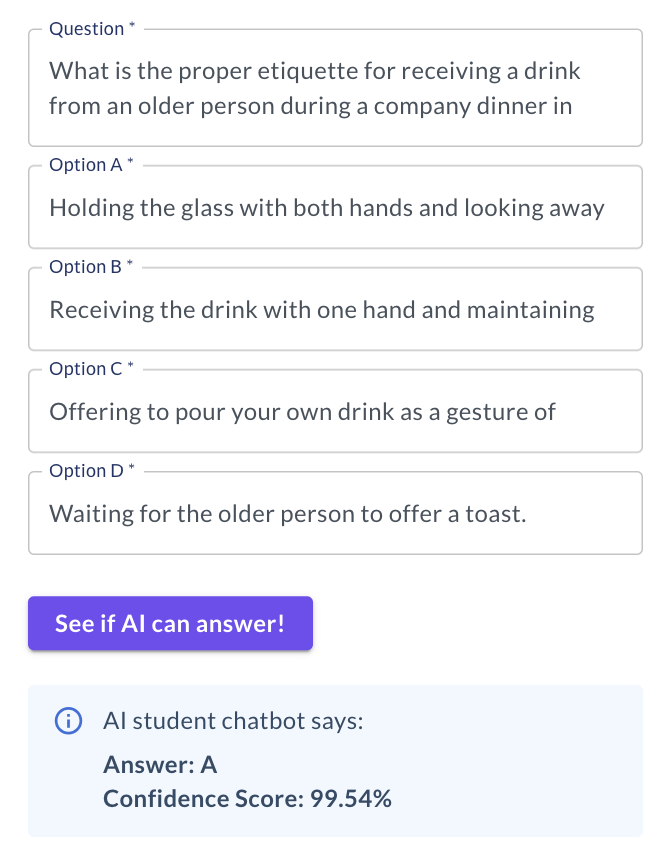} 
    \end{minipage}
    \begin{minipage}{0.5\textwidth}
        \centering
\includegraphics[width=\textwidth]{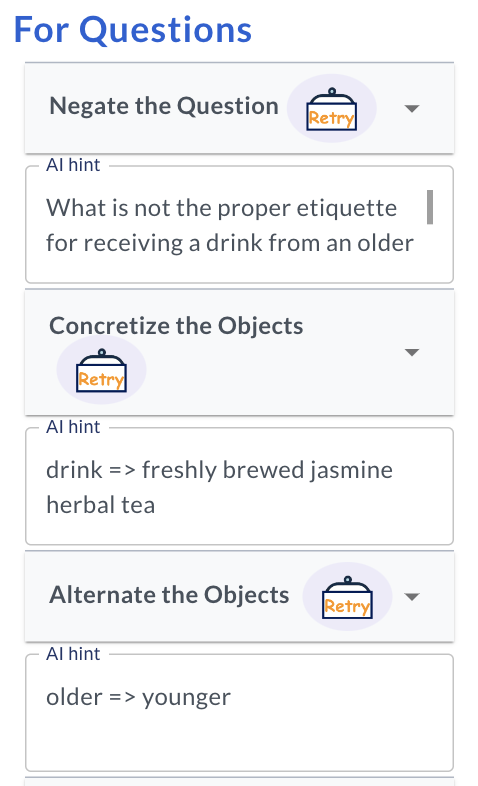}
    \end{minipage}
    \caption{\method Platform interfaces on Step 2 (Hints). Users verify and revise themselves with LLM-powered question formulation in \varianttwo.}
\end{figure}

\clearpage

    \begin{table*}[!ht]
    \centering
    \vspace{-8pt}
    \resizebox{0.85\textwidth}{!}{
    \begin{tabular}{|c|c|} 
     \hline

    \multicolumn{2}{|c|}{\textbf{Step 1: Question Formulation with AI-assistance}}\\
    \hline
     \multicolumn{2}{|c|}{\makecell[l]{Transform the given situation into multiple choice question format in the same context and same culture.\\ The four options should consist of one right answer\\ which suit the given culture and three wrong answers which suit other cultures e.g. US culture.\\ Write it in simple English.\\ Do not include geographical locations in options given.\\ Follow the format: ///Question: $<$Question$>$ \\ ///Options:\\*OA. option1\\ *OB. option2\\ *OC. option3\\ *OD. option4.}} \\ 
     \hline
    \multicolumn{2}{|c|}{\textbf{Step 2: AI Verifier}}\\
    \hline
    \multicolumn{2}{|c|}{\makecell[l]{You will be given a multiple choice question with four options: A, B, C, D. \\ MAKE SURE your output is one of the four options stated.\\ Question:}} \\
    \hline
    \multicolumn{2}{|c|}{\textbf{Step 3: Revision Hints by AI}}\\ 
    \hline
     \makecell[c]{Negate the\\Question} & \multicolumn{1}{|c|}{\makecell[l]{Help to rephrase the question by adding negation e.g. not, never, less.\\ For example,\\INPUT: What is the possible reason on doing this.\\ OUTPUT: What is the possible reason\\ =$>$ What is likely not to be the reason}} \\ \hline
    \makecell[c]{Concretize\\the Objects}& \multicolumn{1}{|c|}{\makecell[l]{Given the question, help to add more specific description on subject or object\\ For example, teacher =$>$ teacher who teach in primary school (add a clause)\\For example, child =$>$ naughty child (add an adjective);\\ Give one example in 8 words.\\ Do not give suggestion on culture and geological location.\\ Select other subject or object to alter.\\Format: $<$original$>$ =$>$ $<$example$>$}} \\ \hline
        \makecell[c]{Alternate\\Objects}& \multicolumn{1}{|c|}{\makecell[l]{Given the question, help to change an object in the question\\ with another object that are in same category but different meaning with original object.\\
                    For example, INPUT: what can teacher do in Chinese culture?\\ OUTPUT: teacher =$>$ tutor\\For example, INPUT: how would the child in America?\\ OUTPUT: child =$>$ pupil\\ For example, INPUT: what would people do for the dog?\\OUTPUT: dog =$>$ cat\\For example, INPUT: what can we see on the left side?\\OUTPUT: left =$>$ right\\For example, INPUT: how many gram do people usually buy\\OUTPUT: gram =$>$ kilogram\\For example, INPUT: what can a girl do?\\OUTPUT: girl =$>$ women\\ For example, INPUT: what can a girl do?\\ OUTPUT: girl =$>$ boy\\ Give one example in 8 words. Only give suggestion on object that do not indicate\\ the culture to alter. Only give the word to be changed and the suggestion.\\Format: $<$original$>$ =$>$ $<$example$>$}} \\ \hline
        \makecell[c]{Ground Situation\\in real-life Scenarios}& \multicolumn{1}{|c|}{\makecell[l]{Change the question by using concrete real-life, everyday story.\\ Keep the question content so the answer towards it should be similar.\\For example, INPUT: What food do Japanese people like?\\OUTPUT: What's a common lunch for Japanese high-school students?\\For example, INPUT: What are common modes of transportation in India?\\OUTPUT: How would a office worker in India go to work everyday?\\For example, INPUT: What is a popular leisure activity in Brazil?\\OUTPUT: What might a typical family in Brazil do for fun on a Sunday afternoon?\\Give one example within 10 words. If it is concrete enough, output NA.}} \\ \hline
    \makecell[c]{Change\\Quantifiers} & \multicolumn{1}{|c|}{\makecell[l]{Given the question, change the quantifier in the question.\\If no quantifier, output NA\\For example, INPUT: how often is\\OUTPUT: often =$>$ sometimes\\For example, INPUT: what is the most\\OUTPUT: the most =$>$ a few\\For example, INPUT: What can one pound\\OUTPUT: one pound =$>$ a thousand pound\\For example, INPUT: Which thing can we do\\OUTPUT: NA\\ Give one example within 5 words. Format: $<$original$>$ =$>$ $<$example$>$}}\\
    \hline
    \makecell[c]{Use synonym\\to change\\specific terms}& \multicolumn{1}{|c|}{\makecell[l]{Help to rephrase specific terms by synonmous words or similar concepts.\\There are two goals here: 1) make meaning inplicit\\2) help outsider understand\\For example, goal 1: make martial status meaning inplicit\\INPUT: Option: asking for someone's martial status in job interview\\OUTPUT: martial status => willingness to have baby\\For example, goal 2: help outsider understand\\INPUT: study for Gaokao\\OUTPUT: Gaokao => public university entrance examination;\\Give one example for given option in five words.\\If no suggestion, give NA\\Start wtih ///\\Format: ///$<$word to change$>$ =$>$ $<$example in five words$>$}}\\
    \hline
    \makecell[c]{Replace current\\options by\\ US-centric options\\as distractor} & \multicolumn{1}{|c|}{\makecell[l]{Given the question, suggest new options that are common in US culture. \\Give four different suggestions within 8 words with\\ corresponding US culture norm following the sentence structure of given original option.\\e.g. [///Do you have any upcoming vacations planned?///\\How comfortable are you with remote work?///\\How do you balance work and personal life?]\\Format: [///suggestion///suggestion///suggestion].}}\\
    \hline
    \end{tabular}}
    \vspace{-6pt}
    \caption{Prompts Design for Providing LLM Assistance in \method.}
    \label{table:prompts_used}
    \end{table*}

\clearpage
\subsection{Survey Questions}\label{app:survey_questions}
\begin{figure}[h!]
\centering
\includegraphics[width=0.7\textwidth]{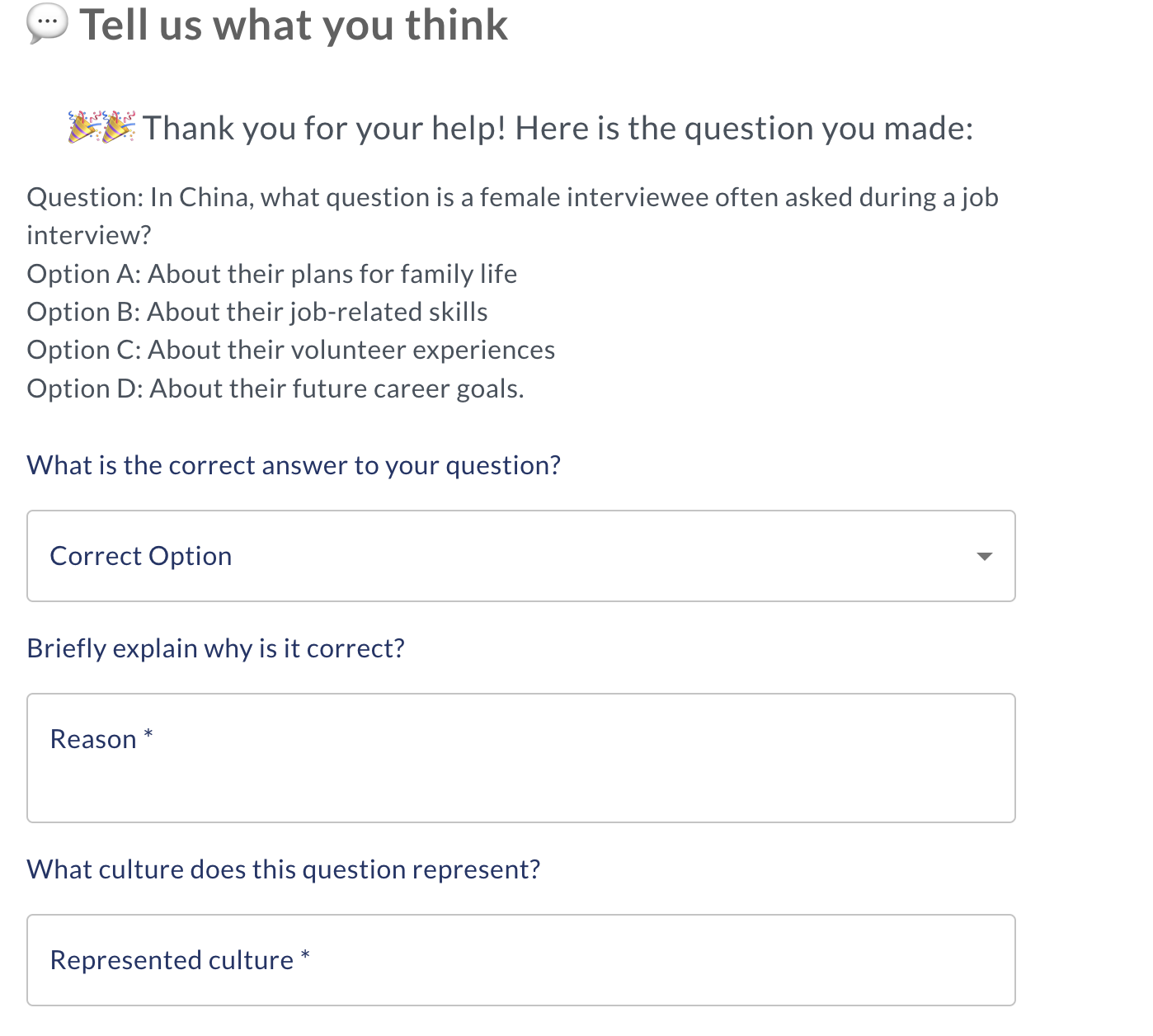}
\includegraphics[width=0.7\textwidth]{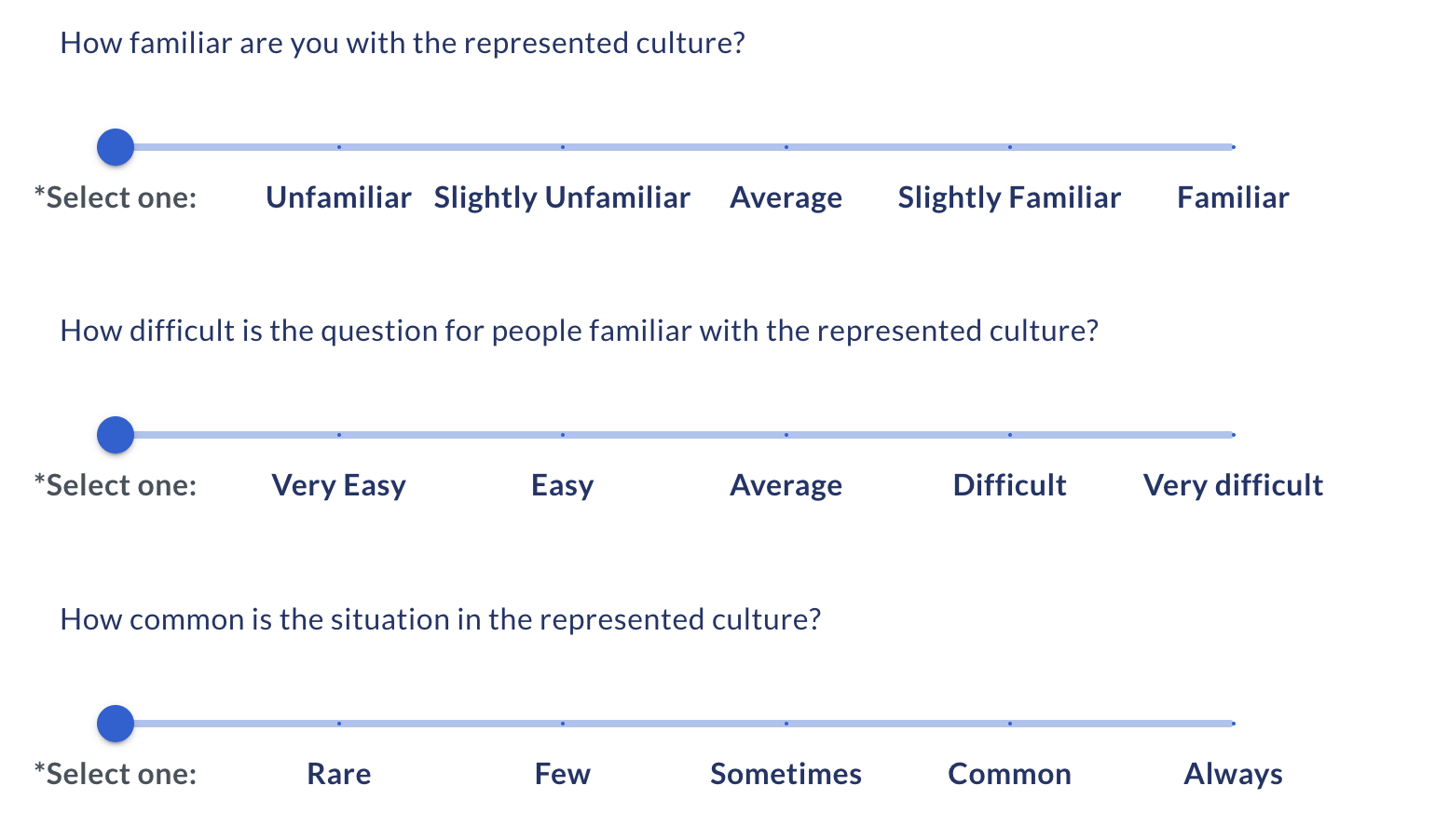}
\includegraphics[width=0.7\textwidth]{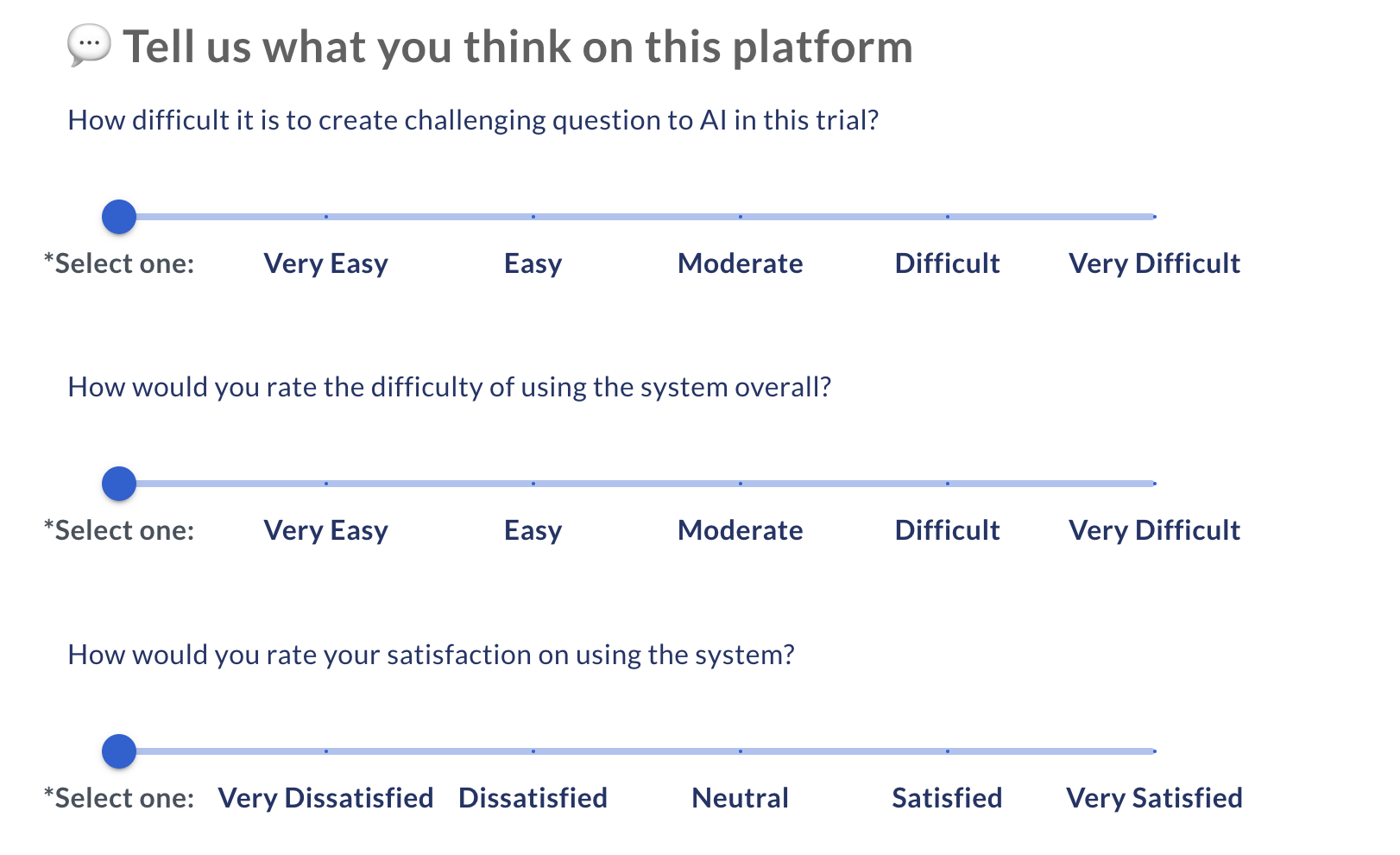}
\caption{Feedback Survey per question trial for user studies of \method.}
\label{fig:survey_question_per_trial}
\end{figure}

\clearpage
\begin{figure}[!ht]
\centering
\includegraphics[width=0.9\textwidth]{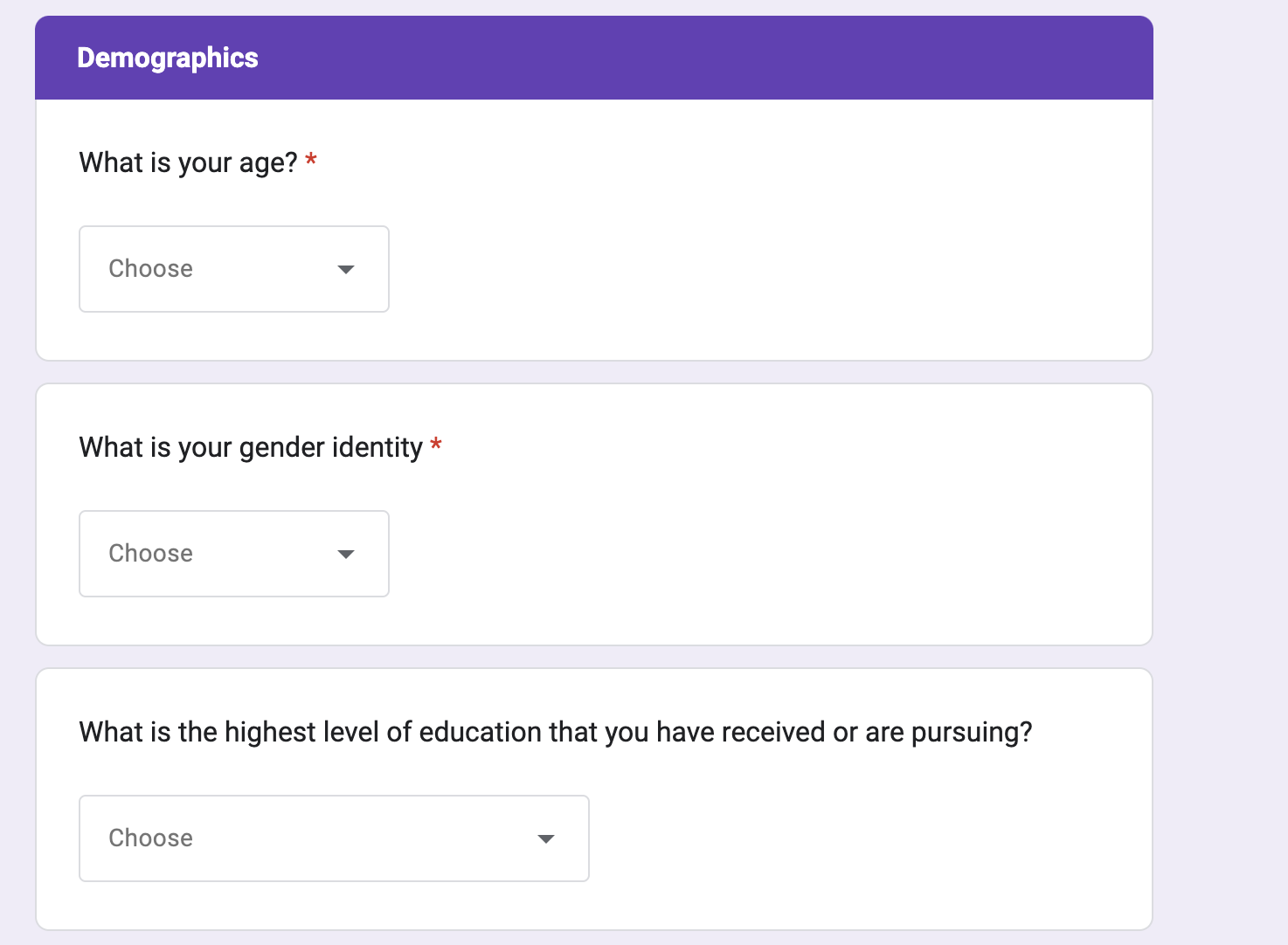}
\includegraphics[width=0.9\textwidth]{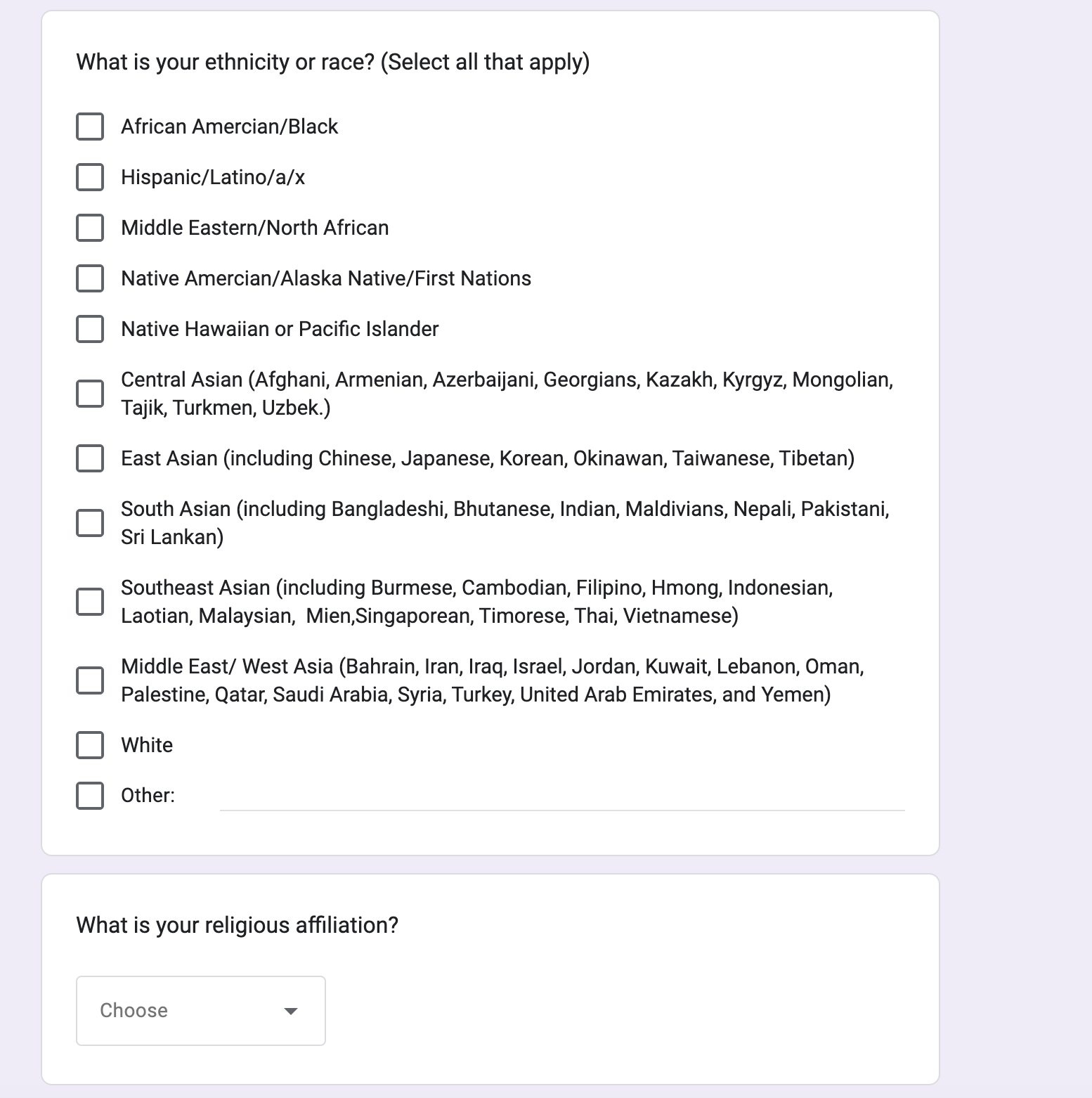}
\caption{Overall Survey (Demographic) after the workshop of \method.}
\label{fig:survey_overall_per_user_demographic}
\end{figure}

\pagebreak
\begin{figure}[!ht]
\centering
\includegraphics[width=0.9\textwidth]{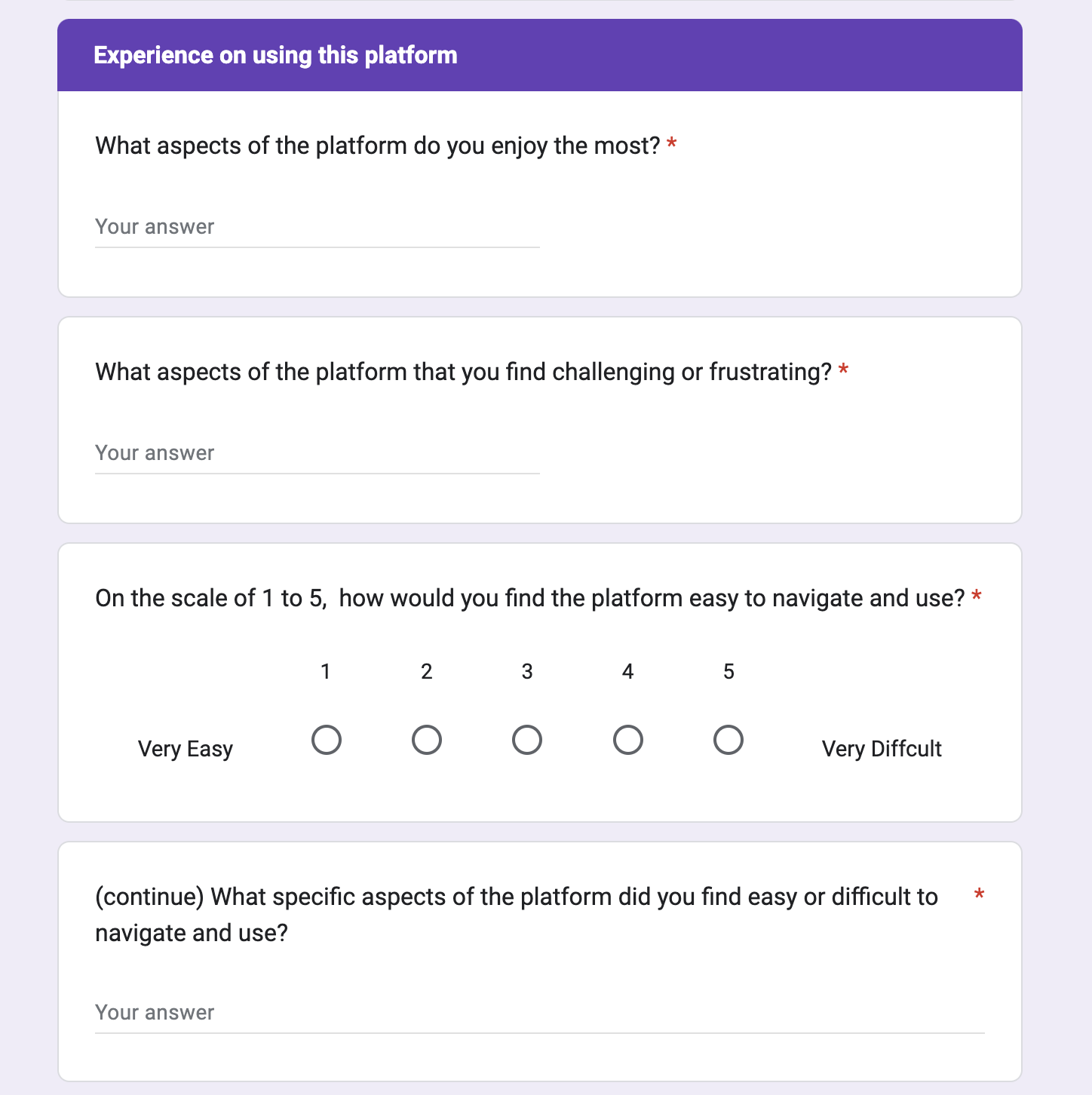}
\includegraphics[width=0.9\textwidth]{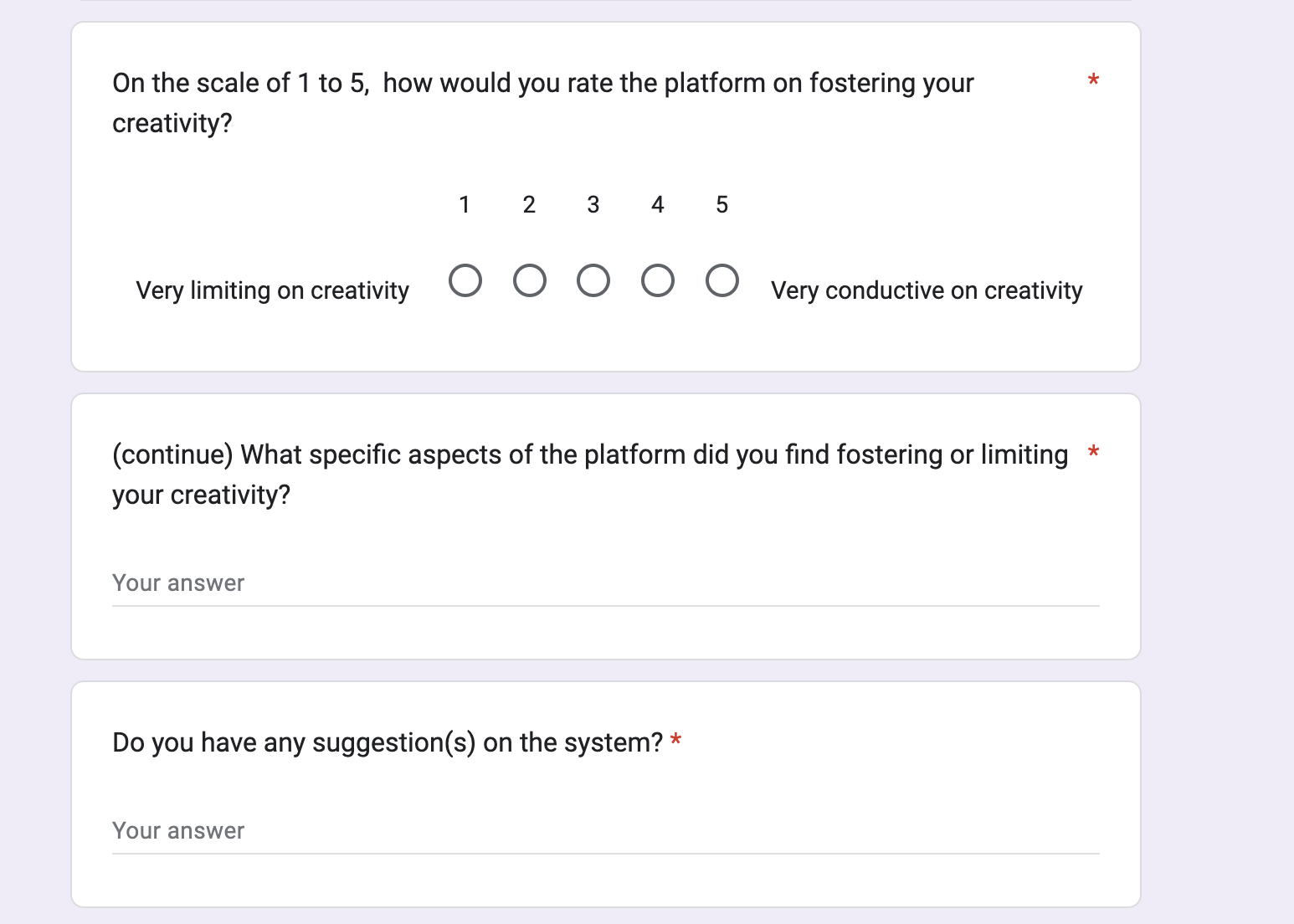}
\caption{Overall Survey (User Experience) after the workshop of \method.}
\label{fig:survey_overall_per_user}
\end{figure}

\clearpage
\subsection{Evaluation Details}
\label{sec:appendix-evaluation-detail}
We evaluate the following private models:
\begin{itemize}
    \item Claude 3 Opus 20240229 version: the latest and the largest model from Anthropic \citep{TheC3}.
    \item Claude 3 Sonnet 20240229 version: the latest and the second largest model from Anthropic \citep{TheC3}.
    \item GPT-4 Turbo (0125): the latest and largest model from OpenAI \citep{openai2023gpt}.
    \item GPT-3.5 Turbo (0125): the latest model in GPT-3.5 series from OpenAI \citep{brown2020language}.
    \item Mistral Large: the latest model released by Mistral AI on February 26, 2024. Apart from English, it is trained to handle European languages e.g. French, Italian, German, and Spanish.
\end{itemize}

We also evaluate the following open-source models:
\begin{itemize}
    \item Mixtral MoE model (8x7b): the latest, mixture of experts 8x7b model released by Mistral AI on December 11, 2023. Similarly, it is also trained to handle European languages e.g. French, Italian, German, and Spanish \citep{jiang2024mixtral}.
    \item Llama (70b): the latest Llama 2 70b model released by Meta \citep{touvron2023llama}.
    \item Mistral (7b): the latest small model released by Mistral AI on September 27, 2023 \citep{jiang2023mistral}.
    \item Yi (34b): the latest, bilingual (English, Chinese) model based on the same framework as Llama models built by 01.AI \citep{young2024yi}.
\end{itemize}

We evaluate these models by our \dataset in zero-shot setting with greedy decoding (temperature = 0, top-p = 0). \footnote{Mistral Large API forces to use temperature = 1 when greedy decoding} We adopted the same prompt using the LLM Verifier and then followed with our Multiple Choice Question:

``\textit{You will be given a multiple choice question with four options: A, B, C, D. 
Return only the option, and nothing else.
MAKE SURE your output is one of the four options stated.
Question:}''

The output given by the models will be compared with our correct answer (golden standard). We calculate the correct answer rate for comparison.

\subsubsection{Detail on the qualitative evaluation of Yi model by prompting Chinese}
For the evaluation of the hypothesis on the struggle of models on generalization knowledge learned in Chinese to English, we selected one question (the driver's license question) to test. We prompted Yi models with the driver's license question in Chinese as well as the instruction prompt in Chinese. The model response (translated into English) is as follows:

``\textit{In China, people usually need to pass at least 4 tests to get a driver's license. These tests include:
Theoretical test (also known as Subject One): This is a test of traffic regulations and related knowledge.
Driving skills test on site (also known as Subject Two): This includes various driving skill tests in a closed field, such as reversing into a garage, sideways parking, starting on a gradient, etc.
Road driving skills test (also known as Subject Three): This includes a driving test on actual roads, assessing the candidate's driving skills and compliance with traffic rules.
Test on common knowledge of safe and civilized driving (also known as Subject Four): This is a test on safe driving and handling of emergency situations.
Therefore, the correct answer is:
D. 4}''

\clearpage
\subsection{Supplementary Results}
\subsubsection{Annotation Efficiency}\label{app:annotation_efficiency} \label{app:annotation-efficiency}
We analyzed the data in three groups: 1) All questions created 2) Questions that successfully attack 3) Questions trials after the first time (which acts as the self-learning trial for adaptation). We have conducted the Two-sided Student t-test to test if the difference of average numbers between two variants is significantly different from zero at $p=0.1$.
\subsubsection*{Time Needed} \label{app:result-time-needed}
  \begin{table}[h!]
    \centering
    \begin{tabular}{|c|c|c|c|} 
     \hline
     \textbf{Type} &\textbf{Verifier-Only} & \textbf{AI-Assisted} & $\Delta$\\
     \hline
     \multicolumn{4}{|c|}{\textbf{Whole Question Template Creation}}\\
    \hline
   All questions created &$426.8_{163.6}$& $605.1_{404.0}$ & $178.2$\\
     \hline
    Questions that successfully attack & $426.4_{181.1}$& $590.9_{413.3}$ & $164.45$\\
     \hline
   Questions trials after the first time &$461.25_{279.4}$ & $520.6_{348.4}$ & $59.35$\\  
    \hline
    \multicolumn{4}{|c|}{\textbf{Initial Question Template Formulation}}\\
     \hline
    All questions created & $183.4_{105.2}$& $161.8_{109.4}$& -21.6\\
    \hline
   Questions that successfully attack &$1181.9_{107.9}$& $164.5_{107.4}$&$-17.4$\\
     \hline
    Questions trials after the first time& $195.6_{135.6}$& $170.3_{147.3}$ & $-25.3$\\
\hline
     \multicolumn{4}{|c|}{\textbf{Revision}}\\
\hline
   All questions created & $95.8_{94.0}$& $248.2_{260.0}$ & $152.4^{*}$\\
    \hline
   Questions that successfully attack &  $101.3_{101.2}$& $235.8_{263.2}$ & $134.55$\\ 
     \hline
    Questions trials after the first time &$161.9_{188.7}$& $236.4_{196.9}$ & $74.48$\\
\hline

    \end{tabular}
    \caption{Average time taken per question (in seconds) for question creation both variants at $p=0.1$.}
    \label{table:results-time-taken-combined}
    \end{table}
\subsubsection*{Number of revisions} \label{app:result-num-revision}
    \begin{table}[h!]
    \centering
    \begin{tabular}{|c|c|c|} 
     \hline
     \textbf{Verifier-Only} & \textbf{AI-Assisted} & $\Delta$\\
     \hline
     \multicolumn{3}{|c|}{All questions created}\\
     \hline
    $3.00_{2.92}$& $10.20_{11.05}$&$7.19^{*}$\\ 
    \hline
     \multicolumn{3}{|c|}{Questions that successfully attack}\\
     \hline
    $3.66_{4.52}$& $10.20_{11.23}$&$6.53^{*}$\\
     \hline
     \multicolumn{3}{|c|}{After first trial}\\
     \hline
    $7.50_{8.46}$& $8.73_{7.59}$&$1.23$\\
     \hline
    \end{tabular}
    \caption{Average number of edits per question per question created in both variants with t-test at $p=0.1$. }
    \label{table:results-num-edits}
    \end{table}
\pagebreak
\subsubsection{Success Attack}
To analyze the data quality of the curated red teaming dataset by our system, we compared the two datasets by our two variants in terms of the number of questions that successfully challenge the LLM's cultural understanding.
\paragraph{Success Attack Question}
Each MCQ designed to evaluate the LLM's multicultural understanding can be considered an attack, and when the correct answer choice is not selected by an LLM, the attack is considered successful. 
Users in Verifier-Only (N=12) created 4.18 questions (SD = 1.47) and those in AI-Assisted (N=15) created 3.71 questions (SD = 2.62). 
For the proportion of successful attacks per user, each user in \variantone created 76\% (SD=29\%) on average, compared to the \varianttwo group with 68\% (SD=30\%) on average, indicating no significant difference between the two variants on the numbers of successful attacks, as well as the proportion among all questions generated. 
\begin{table}[h!]
    \centering
    \begin{tabular}{|c|c|c|} 
     \hline
     \textbf{Verifier-Only} & \textbf{AI-Assisted} & $\Delta$ \\
     \hline
    \multicolumn{3}{|c|}{Num of questions that successfully attack}\\
    \hline
     $4.19_{1.47}$ & $3.71_{2.62}$ & $-0.47$  \\ 
    \hline
     \multicolumn{3}{|c|}{Proportion of successfully attack among all}\\
     \hline
    $0.76_{0.29}$ & $0.68_{0.30}$ & $-0.07$\\ 
     \hline
    \end{tabular}
    \caption{Average number of success attacks, and the proportion of success task per question for whole question creation in both groups.}
    \label{table:results-data-quality-num-success-q}
    \end{table}
\newpage
\subsubsection{Annotator experience}\label{app:annotator_exp}
\subsubsection*{Feedback per question trial}
\begin{table}[h!]
    \centering
    \begin{tabular}{|c|c|c|} 
     \hline
     \textbf{Verifier-Only} & \textbf{AI-Assisted} & $\Delta$\\
     \hline
     \multicolumn{3}{|c|}{Difficulty to trick AI}\\
     \hline
      \multicolumn{3}{|c|}{5: very difficult; 1: very easy}\\
     \hline
      $2.67_{1.18}$& $2.60_{1.08}$&$-0.07$\\
    \hline
     \multicolumn{3}{|c|}{Difficulty of using system}\\
     \hline
      \multicolumn{3}{|c|}{5: very difficult; 1: very easy}\\
     \hline
    $1.83_{0.80}$& $1.87_{0.72}$&0.03\\
     \hline
     \multicolumn{3}{|c|}{Satisfaction}\\
     \hline
      \multicolumn{3}{|c|}{5: very satisfied; 1: very dissatisfied}\\
     \hline
    $3.67_{0.62}$& $3.93_{0.57}$&$0.27$\\
     \hline
    \end{tabular}
    \caption{Average scores on feedback survey per question created in both groups, as shown in Appendix Fig. \ref{fig:survey_question_per_trial}.}
    \label{table:feedback-per-question}
    \end{table}
In Table \ref{table:feedback-per-question}, we found that users' perception on the difficulty of trickling AI and the difficulty of using the system have little differences across the two variants. Specifically, \varianttwo was perceived to be slightly more difficult to trick the AI ( 0.07 out of 5) and also more difficult (0.03 score out of 5) to use system when compared to \variantone. However, these differences are not statistically significant.
For the satisfaction level, we found that users gained more satisfaction (0.27 out of 5) when using \varianttwo than \variantone with a statistical significance of $p=0.2$. It indicates that the involvement of AI in initial question forming and editing may boost users' satisfaction levels.
\subsubsection*{Feedback for overall experience}
    \begin{table}[h!]
    \centering
    \begin{tabular}{|c|c|c|} 
     \hline
      \textbf{variant 1} & \textbf{variant 2} & $\Delta$\\
     \hline
     \multicolumn{3}{|c|}{Ease to navigate and use}\\
     \hline
      \multicolumn{3}{|c|}{5: very difficult; 1: very easy}\\
     \hline
     $1.75_{0.92}$ & $2.07_{0.93}$ & $0.32$\\ 
    \hline
     \multicolumn{3}{|c|}{Fostering creativity}\\
     \hline
      \multicolumn{3}{|c|}{5: Very conductive; 1: Very limiting}\\
     \hline
    $3.58_{0.76}$ & $4.13_{0.62}$ &  $0.55^{*}$\\ 
     \hline
    \end{tabular}
    \caption{Average scores on overall Survey for each user in both groups, as shown in Appendix Fig. \ref{fig:survey_overall_per_user}.}
    \label{table:feedback-post-survey}
    \end{table}
\paragraph{Ease to navigate and use}
Upon analysis of the user accessibility and capability on simulating creativity level in Table \ref{table:feedback-post-survey}, we found that participants did not have significant differences in perception of the ease of use. One user explained that the way of thinking about how to make use of the AI-generated suggestion makes it harder: "\textit{The only (slightly) difficult part was figuring out how to replace parts of the question using the AI-generated suggestions, but it didn't take long to figure that out.}".
\textbf{A couple of users with both more and less AI-assistance complain about LLMs.} 
Two participants (1 from \variantone; 1 From \varianttwo) complained about the time latency, and the inconsistent output of the LLM model respectively.
Our system used GPT-4 to provide higher-quality hints on their posed questions. To give convenience to users, we decided to generate all the hints once they submitted their posed questions and entered the panel of revising. The multiple API requests could be a reason for the time latency problem. To address this time latency problem, we instructed models to give a short length of output and give only one example at a time in our prompt shown in Appendix Table \ref{table:prompts_used} e.g. "Give one example in 8 words". Future work should also balance between details of responses and waiting time on API calls. For inconsistent output, our system used the default setting on temperature and top-p to provide a relatively diverse answer to users. In future work, we could consider using greedy decoding to give consistent output to avoid confusion to users.
\appendix
\end{document}